\documentclass{article}
\usepackage[nonatbib, preprint]{neurips_2026}
\usepackage{graphicx} %
\usepackage{amsmath}

\usepackage{booktabs} %
\usepackage{longtable} %
\usepackage{multirow} %
\usepackage{array} %
\usepackage{paralist} %
\usepackage{verbatim} %
\usepackage{subfig} %
\usepackage{tikz,pgfplots}
\pgfplotsset{compat=1.18}
\usetikzlibrary{arrows.meta,positioning,calc}

\definecolor{darkblue}{RGB}{0,70,160}

\usepackage[labelfont=bf, format=plain]{caption}
\usepackage{mathtools}
\usepackage{amssymb}
\usepackage{wrapfig}
\usepackage{amsmath}

\usepackage[
  style=numeric-comp,
  sorting=none,
  url=false,
  maxbibnames=99
  ]
  {biblatex}
\AtEveryBibitem{\clearlist{language}}

\usepackage[colorlinks=true, linkcolor=darkblue, citecolor=darkblue, urlcolor=darkblue]{hyperref}

\renewcommand{\vec}{\mathbf}
\newcommand{\T}{\top}

\title{Teacher Geometry Shapes Learnability in Teacher-Student Networks}

\author{%
  Kai J. Sandbrink\thanks{kai.sandbrink@lmh.ox.ac.uk} \\
  School of Computer and Communication Sciences and School of Life Sciences, EPFL\\
  Department of Experimental Psychology, University of Oxford
  \And
  Flavio Martinelli \\
  School of Computer and Communication Sciences and School of Life Sciences, EPFL\\
  \AND
  Alexander van Meegen \\
  School of Computer and Communication Sciences and School of Life Sciences, EPFL\\
  Department of Physics, RWTH Aachen \\
  \And
  Wulfram Gerstner\thanks{Co-senior authors.} \\
  School of Computer and Communication Sciences and School of Life Sciences, EPFL\\
  \And
  Johanni Brea\footnotemark[\value{footnote}] \\
  School of Computer and Communication Sciences and School of Life Sciences, EPFL\\
}

\begin{document}

\maketitle

\begin{abstract}
    Teacher-student systems, in which a teacher neural network generates training labels so that a student neural network can learn to implement the same function, are widely used as an abstract setting to study learning. However, the structure of the teachers is often overlooked by assuming randomly-generated, normally-distributed parameters. This hides substantial variation in how learnable different teachers are. We formalize learnability as the success rate of converging to the global minimum, as a function of overparameterization, learning algorithm, student initialization distribution, and teacher geometry. We both identify an easy distribution that maximizes node dissimilarity and a hard distribution that minimizes it, and show that these two distributions induce markedly different success rates across a large range of settings and for different activation functions. To explain the gap, we study the loss landscape of small neural networks that contain two distinct kinds of suboptimal local minima, out-of-bounds (OOB) minima at the edge of the data distribution and interior minima within. Assuming infinite data and a fast readout layer, we analytically reduce the loss landscape of small networks to two dimensions, showing that the region of attraction of interior minima changes as a function of teacher structure. In larger networks, maximally dissimilar teachers induce more interior minima, while minimally dissimilar teachers induce more OOB minima. Motivated by these analyses, we show that differentially increasing the learning rate of the readout layer and decreasing the learning rate of the inner biases increases success rates. These findings provide an important step in narrowing the gap between the study of teacher-student networks and more structured functions that arise in practice.
\end{abstract}

\section{Introduction}

\label{sec:intro}

How well a function can be learned by a neural network, also known as its learnability, is strongly determined by its expressibility~\cite{yarotsky2017, montufar.etal2014, arora.etal2018}: how many units, and what architecture, does a neural network need to express it? If a neural network does not have sufficient capacity to represent a function, it will never be able to learn it. Linear functions and simple boolean functions AND and OR, which can all be represented by a single node, are the simplest. Famously, XOR requires two nodes and two layers to be solved successfully~\cite{minsky.papert1969, rumelhart.etal1986}. In higher dimensions, parity on multiple bits places requirements on architecture and input data distribution to be learned successfully~\cite{daniely.malach2020, shoshani.shamir2025, barak.etal2022}. Functions containing periodic components are hard for neural networks to approximate across common activation functions, none of which contain a periodic element~\cite{ziyin.etal2020}, and therefore require a large number of neurons. 
However, this is not the complete picture. Even when a function can be efficiently implemented by a neural network, local optimization methods such as gradient descent may still fail to find the solution by converging to suboptimal local minima~\cite{safran2018spurious, shalev-shwartz.etal2017}. 

We therefore ask: Apart from expressibility, what factors influence the learnability of a function? 
To make this question precise, we make use of the teacher-student framework in which a student is trained to match the outputs generated by another network~\cite{saad.solla1995, saad.solla1995a, seung.etal1992}.  This makes a minimal parameterization known by construction.

Empirically, the rate of convergence to the global minimum is highly dependent on the exact configuration of the teacher networks, as demonstrated by ``easy teachers'' that are learnable by a student with the same number of nodes and ``hard teachers'' that are rarely matched even by student networks twice as wide as the teacher network \cite{martinelli.etal2024}. 
Teachers in many theoretical studies are drawn with normally distributed weights~\cite{seung.etal1992, saad.solla1995, advani.etal2020, aubin2018committee, goldt.etal2019}, or assume simplified structures on weights or data (e.g. diagonal~\cite{pesme2023saddle}, orthogonal~\cite{boursier2022gradient});
these modeling choices do not take into account differences in the difficulty between teachers. 
A complete picture of what factors determine the probability of a student network reaching zero loss is missing.

In this paper, we systematically study the sources of variation in difficulty of teacher-student systems from the perspective of loss landscapes. For first-order optimization methods such as gradient descent, the critical points and their regions of attraction are of particular interest as they can lead student networks to converge above the global minimum~\cite{safran2018spurious}. The loss landscape with its critical points is purely determined by the teacher network (and the training data). However, the regions of attraction depend also on the flow fields induced by the optimizer dynamics \cite{surace.etal2020}.
Therefore, two teacher networks of the same size can induce loss landscapes of different levels of difficulty either by increasing or decreasing the number of suboptimal local minima; or by changing the size of their basins of attraction.
In this paper, we investigate these factors through the study of three teacher distributions with highly different success rates, through an analytical reduction to study the loss landscape of small networks and through a number of controlled experiments to identify specific factors of variance.

The paper makes the following contributions:

\begin{itemize}
    \item We introduce and formalize teacher-student learnability as a function of teacher distribution, initialization, data, overparameterization, and optimizer.
    \item We discover that the similarity between teacher node weights strongly negatively correlates with the convergence probability of students. We define three distributions of teachers with increasing degree of learnability: minimally dissimilar, standard, and maximally dissimilar.
    \item We characterize the loss landscape in ReLU networks analytically to study how local minima are influenced by the orientations, signs, and biases of teacher and student nodes.
    \item We describe the minima reached by students in the different teacher distributions, showing that neurons are more likely to be driven out-of-bounds for minimally dissimilar and standard distributions than for the maximally dissimilar teacher distribution.
    \item We study the impact of the optimizer on the learning algorithm and show in particular that increasing the learning rate of the readout weights or decreasing that of the inner biases substantially improves success rates across all distributions.
\end{itemize}

\begin{figure}
   \centering
   \includegraphics[width=\textwidth]{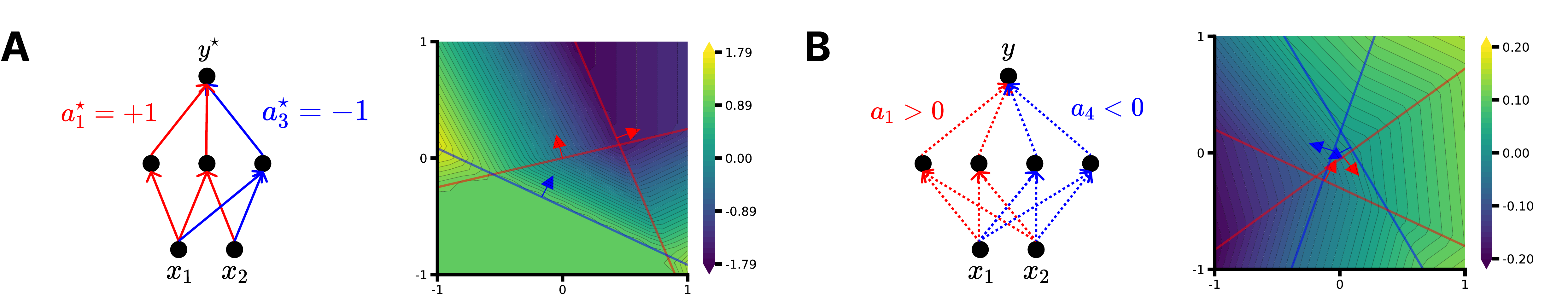} 
\caption{(\textbf{A}) A teacher network with input dimension $D=2$ and $M=3$ hidden ReLU neurons. The color of the lines indicates whether a node has a positive (\textit{red}) or negative (\textit{blue}) readout weight. The color in the contour plot indicates the network output as a function of the input $\vec{x}$; the arrows indicate the non-zero side of the ReLU. (\textbf{B}) Same as (A), except showing an overparameterized student network at initialization with $N=4$ hidden neurons.}
\label{fig:teacher-student}
\end{figure}

\section{Teacher-student learnability}

\subsection{Setting}

We consider a standard teacher-student framework~\cite{seung.etal1992} with networks containing one hidden layer. 
Let $D \in \mathbb{N}$ denote the input dimensionality, and $X \sim \mathcal{X}$ a set of data points $\vec{x}_i \in \mathbb{R}^D$ drawn from the input distribution $\mathcal{X}$.
In this setting, a \textit{teacher network} is defined through embedding weights $\vec{W}^\star\in \mathbb{R}^{M \times D}$, readout weights $\vec{a}^\star \in \mathbb{R}^{M}$, inner biases $\vec{b}^\star \in \mathbb{R}^M$, and an outer bias $c^\star \in \mathbb{R}$, where $M$ is the number of neurons, and a teacher activation function  $\sigma^\star$ (e.g.\ ReLU, softplus, or tanh; see Figure~\ref{fig:teacher-student}A for an example).
The weights of the teacher are drawn according to a distribution $\mathcal{T}$. For each $\vec{x}_i \in X$, the teacher then generates output labels
\begin{equation}
    y^\star_i = \vec{a}^{\star \T} \, \sigma ( \vec{W}^\star \vec{x}_i + \vec{b}^\star) + c^\star \, .
\end{equation}
The teacher-student problem consists of training a \textit{student network} to match the outputs of the teacher. The student is defined by the embedding weights $\vec{W} \in \mathbb{R}^{N \times D}$, inner biases $\vec{b} \in \mathbb{R}^{N}$, readout weights $\vec{a} \in \mathbb{R}^{N}$, and outer bias $c \in \mathbb{R}$, where $N$ is the number of neurons in the student. The parameters are drawn from an initial distribution $\mathcal{I}$ and change during learning (Figure~\ref{fig:teacher-student}B). %
Given a student activation function $\sigma$, the output of the student network is
$
    y_i = \vec{a}^\T \, \sigma (\vec{W} \vec{x}_i + \vec{b}) + c\, .
$

The overparameterization ratio is $\rho = N / M$. The parameters of the student neural network are optimized on a large dataset of $K$ input-output pairs generated by the teacher. Using a given optimizer, the goal is to minimize the loss function
\begin{equation}
\mathcal{L}(\vec{W}, \vec b, \vec a, c) = \frac{1}{2K} \sum_{i=1}^K (y_i^\star - y_i)^2
\end{equation}
We optimize this loss with gradient flow (see Appendix~\ref{app:training} for details of the training procedure).

\subsection{Points of convergence}\label{sec:convergence_points}

For units with ReLU activation functions~\cite{glorot.etal2011}, the signed distance from the origin (in the direction of its weight vector $\vec{w}$) describes the position of its \textit{kink} $k = - \frac{b}{||w||}$. We describe a neuron as \textit{in-bounds} if there is at least one datapoint on either side of its kink, described by $\min_i \vec{w} \cdot \vec{x}_i < k < \max_i \vec{w} \cdot \vec{x}_i$; if not, the neuron is \textit{out-of-bounds}. This allows us to distinguish the following points to which the system can converge for a dataset with a finite number of data points:
\begin{itemize}
    \item \textbf{Global minima $\mathcal L = 0$}: Each teacher node is matched by at least one student node (%
    potentially with symmetries~\cite{simsek.etal2021,martinelli.etal2024}).
    \item \textbf{Out of bounds (OOB) local minima $\mathcal L > 0$}: A student fails to match a teacher because either one or more nodes are out of bounds pointing outwards (dead neurons ~\cite{lu.etal2020}), out of bounds pointing inwards (linear neurons~\cite{dobler.lemmerich2025}), or both. 
    \item \textbf{Interior local minima $\mathcal L > 0$}: A student fails even though all nodes are in-bounds, but have converged to a suboptimal local minimum.
\end{itemize}

As the probability of a network to reach a saddle point exactly is vanishingly small, it will reach a point belonging to one of the three classes above if trained to convergence. Linear neurons can correspond to true local minima if they have the wrong readout sign for the residual (Appendix~\ref{app:boundary-dynamics}).

\subsection{Learnability}

We define teacher-student learnability for a given teacher network $T$, overparameterization ratio $\rho = N/M$, and optimizer as the expected conditional probability of reaching the global minimum
\newcommand{\success}[1]{\mathbb P(\text{global minimum})}
\begin{equation}
\success{T, \rho, \text{optimizer}, \mathcal I, \mathcal X} = \mathbb E_{I \sim \mathcal{I}, X \sim \mathcal{X}}\left[\mathbb I(\mathcal L = 0\ |\ I, X, T, \rho, \text{optimizer})\right]\, ,
\end{equation}
 for student initializations $I$ drawn from $\mathcal I$ and training data $X$ drawn from $\mathcal X$, where $\mathbb I(\text{true}) = 1$ is the indicator function.
We estimate this probability empirically by training multiple student networks (each with a different random initialization seed) on the same teacher, and measuring the fraction that converge to a loss below $10^{-18}$ (see Figure~\ref{sfig:loss-threshold}). For each teacher, we fit student networks with independent seeds, yielding a binomial estimate of the learnability for that teacher.

\section{Comparing three different teacher distributions}

\subsection{Defining maximally dissimilar, standard, and minimally dissimilar distributions of teachers}

\begin{figure}[t]
    \includegraphics[width=\textwidth]{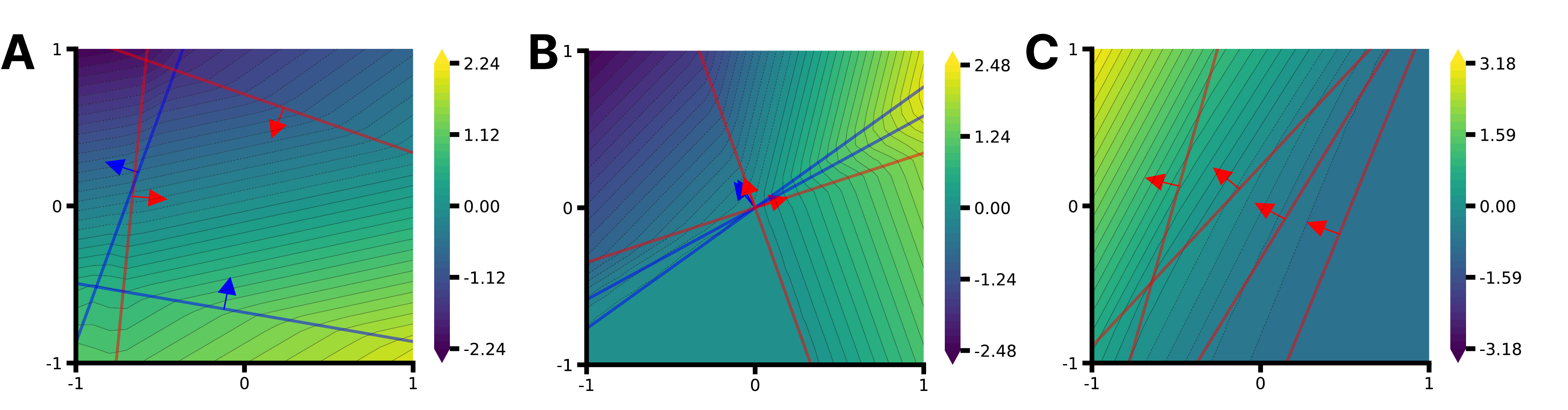}
    \caption{\textbf{Example teachers with four nodes from the three different distributions.} (\textbf{A}) A teacher drawn from the maximally-dissimilar distribution. The lines correspond to the individual nodes' hyperplanes, i.e. where $\vec{w}^\T \vec{x} + b=0$. The arrow indicates the region where the node has a positive output. The color of the arrow and the hyperplane indicates whether the readout sign $a$ of the node is positive or negative. (\textbf{B}) Same as (A), but for teachers from the standard distribution. (\textbf{C}) Same as (B), but for teachers from the minimally dissimilar distribution.}
    \label{fig:distributions}
\end{figure}

To investigate the features of a teacher network that influence student trainability, we consider teacher networks drawn from different distributions.
Under the standard distribution, common in studies of teacher-student setups~\cite{advani.etal2020, goldt.etal2020, camilli.etal2026, goldt.etal2021, seung.etal1992, saad.solla1995, goldt.etal2019}, weights are drawn independently from a normal distribution and biases are zero. But the distribution of weights in trained neural networks differs substantially from the standard parameter initialization: The learned weight structure is richer~\cite{martin.mahoney2021}. To elucidate how teacher geometry shapes learnability, we study two additional teacher distributions. To introduce these distributions, we draw the embedding weight vector $\vec{w}_i$ of each node $i$ in the hidden layer as well as its readout weight $a_i$ as follows; Figure~\ref{fig:distributions} shows example teachers drawn from each of the three distributions.

\begin{itemize}
    \item \textbf{Maximally dissimilar}: 
    Weight vectors $\vec{w}_i$  are chosen to have directions as dissimilar from each other as possible, while avoiding additional symmetries from exactly opposing directions (see Appendix~\ref{app:teacher-construction} for procedure). The readout weights are randomly chosen so that exactly half the nodes have readout weight $+1$ and the other half $-1$. %
    \item \textbf{Standard}: Weight vectors $\vec{w}_i$ of all teacher nodes have weights drawn from a normal distribution and zero bias (so their hyperplanes pass through the origin). The readout weights are sampled from $\{1, -1\}$ for each node.
    \item \textbf{Minimally dissimilar}: Weight vectors $\vec{w}_i$ of all teacher nodes are chosen to  point in the same ``cone" of angle $\pi/8$. All nodes have the same readout weight $a_i=1$. Biases are uniformly spaced between $[-0.5, 0.5]$ and assigned randomly to the different nodes. %
\end{itemize}

This choice is motivated by the fact that maximally and minimally dissimilar distributions have larger and smaller expected initial similarity between student and teacher nodes (see Appendix~\ref{app:res-proof}), given by the quantity
\begin{align}
    \label{eqn:initial-overlap}
    S(\mathcal T, \mathcal I) &= \mathbb E_{T\sim\mathcal T, I\sim\mathcal I} \left[
    \max_{p \in P_N} \sum_{j=1}^M a_{p(j)}a_j^\star \vec{w}_{p(j)}^\T \vec{w}_j^\star
    \right]
\end{align}
\noindent where $P_N$ denotes the symmetric group on $N\geq M$ elements, i.e.\ the set of all permutations of student nodes. All teachers have weights that are normalized to magnitude 1 for each node. Similar overlap terms play an important role in the learning dynamics of teacher-student settings~\cite{goldt.etal2019, saad.solla1995, simsek.etal2024}. When we calculate $S(\mathcal{T}, \mathcal{I})$ for $\rho = 1$, we exclude teacher-student systems where $a_{p(j)} a_j^\star < 0$ and $\vec{w}_{p(j)}^\T \vec{w}_j^\star < 0$; see Appendix~\ref{app:meth-overlap}.

\subsection{The different distributions lead to markedly different success rates}

\begin{figure}\includegraphics[width=\textwidth]{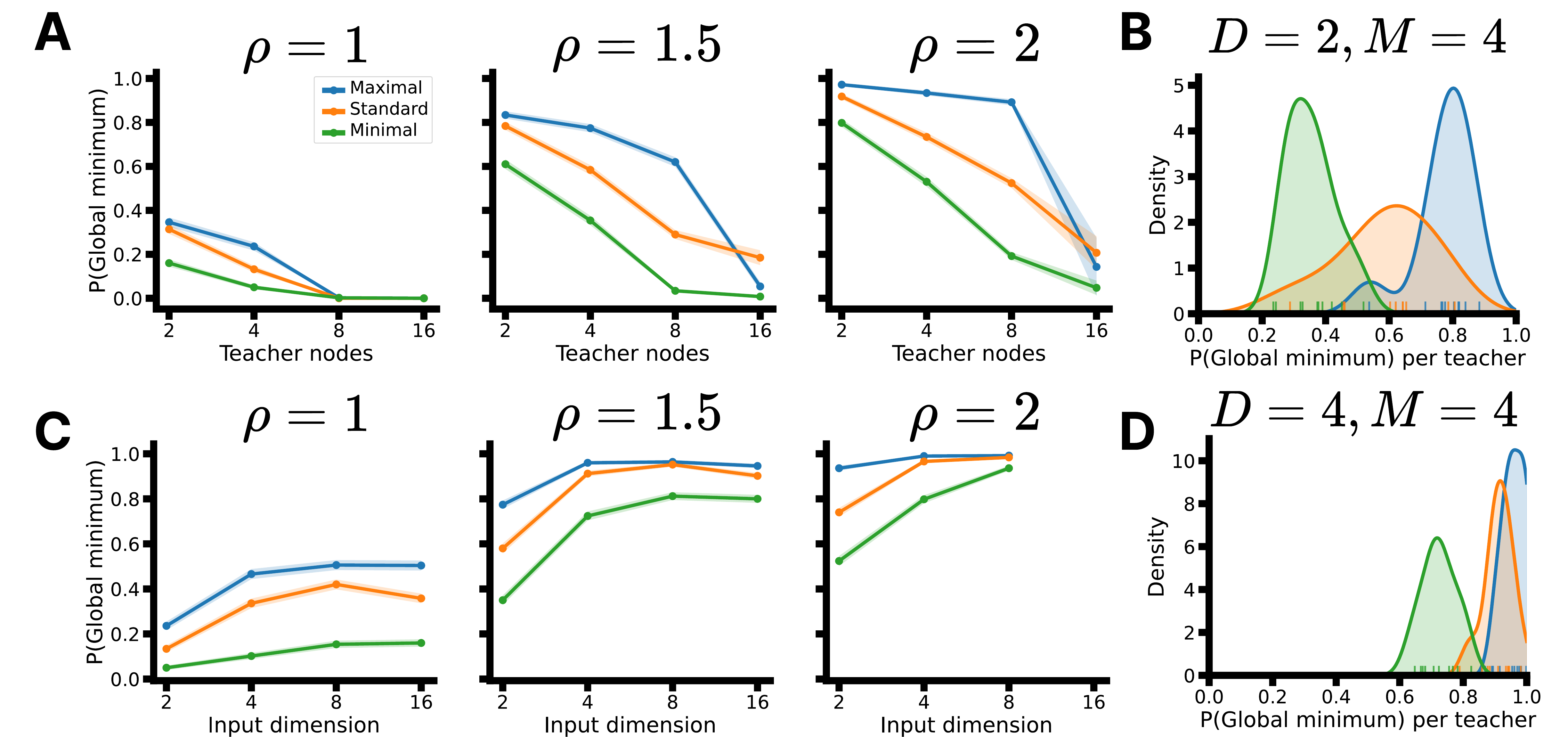}
    \caption{\textbf{The probability of student networks reaching the global minimum depends on the teacher geometry.} (\textbf{A}) Fraction of students reaching the global minimum as the number of teacher nodes are increased for $D=2$. The shaded region indicates the standard error of the binomial proportion.  (\textbf{B}) Frequency of students with $\rho=1$ reaching the global minimum for teachers with $D=2$ and $M=4$, smoothed using Gaussian kernels. (\textbf{C}) Same as (A), but for increasing input dimensionality $D$ and for fixed number of teacher nodes $M=4$. (\textbf{D}) Same as (B), but for teachers with $D=4$ and $M=4$.}
    \label{fig:compare-losses}
\end{figure}

We vary the width of the teacher network $M \in \{2, 4, 8, 16\}$ and the input dimensionality $D \in \{2, 4, 8, 16\}$, changing one parameter at a time while keeping the other constant at $M=4$ or $D=2$. Fitting the student networks results in markedly different levels of convergence to the global minimum. We see that, as the number of teacher nodes increases, the proportion of successful fits decreases (Figure~\ref{fig:compare-losses}A); conversely, as the input dimensionality increases, fitting becomes easier (Figure~\ref{fig:compare-losses}C). In almost all simulations across these settings, the teacher distribution with maximally dissimilar nodes has the highest success rates; then, the standard distribution, and finally, the minimally dissimilar distribution. Within each distribution, individual teachers span a spectrum of difficulty (see Figure~\ref{fig:compare-losses}B,D).

To ensure that the findings generalize beyond the ReLU activation function, we run the same experiment for both the softplus activation function (Figure~\ref{sfig:softplus}) and tanh (Figure~\ref{sfig:tanh}), reaching similar results in both cases. 

\section{Loss landscapes of one- and two-node ReLU systems}
\label{sec:loss-landscapes}

What explains the differences between the different teacher distributions? We consider this question from the perspective of the loss landscape and through targeted experiments in small networks.

\subsection{Single-neuron teacher-student system}

\label{sec:single-neuron}

We can visualize single-node teacher-student systems for dimensions $D>1$ by assuming infinite Gaussian data and a particular optimizer, which sets the linear readout weights always to their optimal value (see Appendix~\ref{app:analytical-reduction}). This has different dynamics than gradient flow, but is easily implementable in real settings, because linear optimization is convex. 
With the change of variables $z=\vec{w}^\T\vec{x}/\|\vec{w}\|$ and $z^\star=\vec{w}^{\star\T}\vec{x}/\|\vec{w}^\star\|$,
and using the homogeneity of ReLU, the infinite data loss landscape can be rewritten as a function of the angle between student and teacher  weights, $\cos\theta=\frac{\vec{w}^\T\vec{w}^\star}{\|\vec{w}\|\,\|\vec{w}^\star\|}$, and the kink, $k=-b/\|\vec{w}\|$, which represents the signed distance between the origin and the hyperplane of the neuron:
\begin{equation}
\label{eq:single-neuron-loss}
\frac{\mathcal{L}(\theta,k)}{\mathrm{const.}}
=
1
-
\frac{
\left\langle \sigma(z-k)\,\sigma(z^\star-k^\star) \right\rangle_{z,z_\star}^{2}
}{
\left\langle \sigma(z-k)^2 \right\rangle_{z}
\left\langle \sigma(z^\star-k^\star)^2 \right\rangle_{z^\star}
}
.\end{equation}
\noindent 
The dependency on $\theta$ enters through the correlation $\langle z, z^\star \rangle = \cos \theta$. Due to the Gaussian data assumption $z,z^\star$ are bivariate Gaussian with zero mean and unit variance. The Gaussian expectations $\langle \sigma (\cdot) \sigma(\cdot) \rangle$ can be computed analytically~\cite{martinelli2025flat}.

\begin{figure}
    \includegraphics[width=\textwidth]{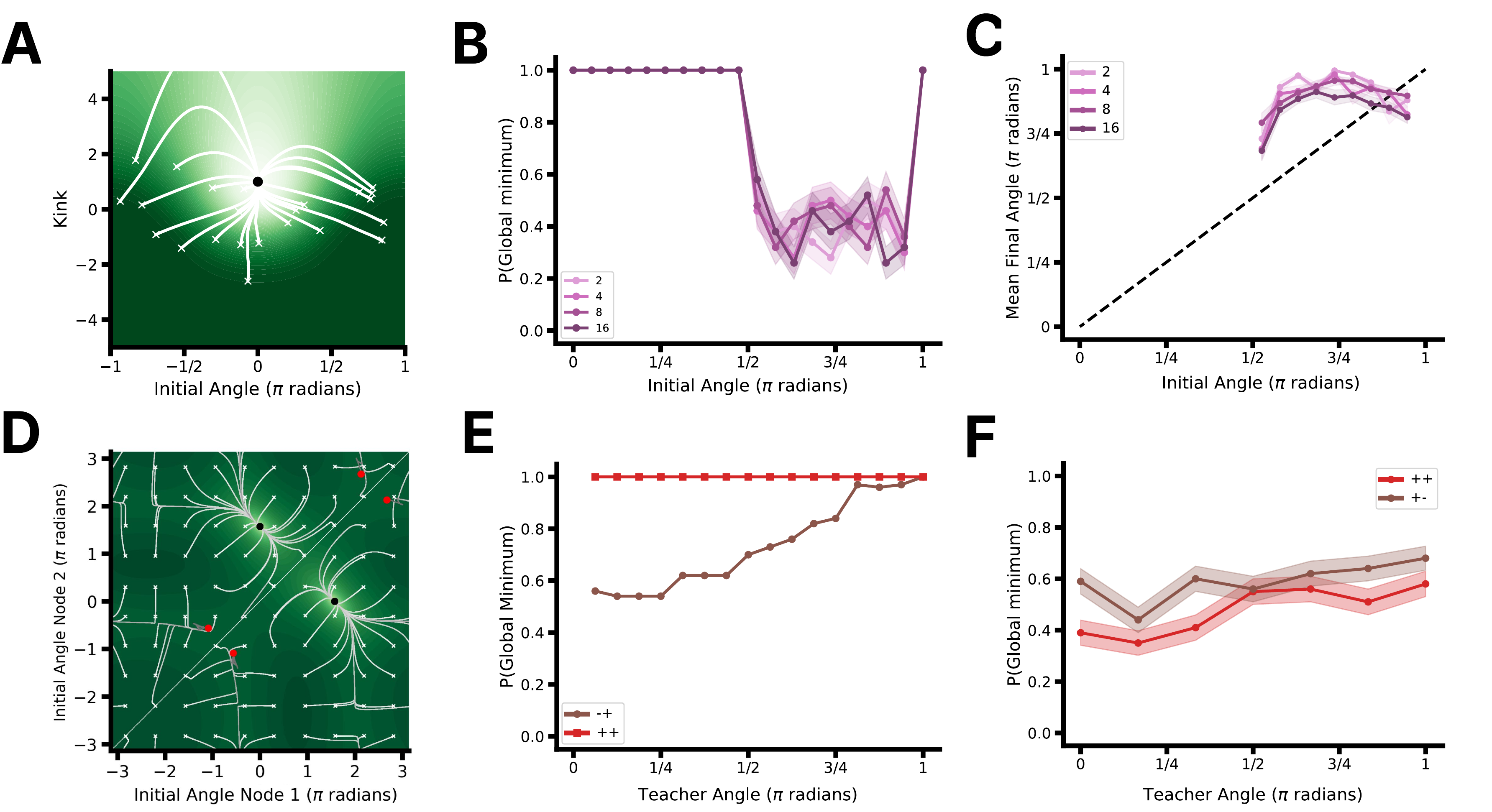}
    \caption{\textbf{Importance of initial similarity between student and teacher neuron.} (\textbf{A}) Loss landscape of a single-neuron teacher-student system assuming infinite data, fast readout weights, and $k^\star=1$. The loss varies from low (\textit{light green}) to high (\textit{dark green}), with individual trajectories (\textit{white lines}) reaching the global minimum (black dot). Note the excursion to large values of $k$, for the most top-left trajectory: without infinite data it would not have returned; instead it would be stuck in an OOB minimum. (\textbf{B}) Success rates as a function of different initial angles between student and teacher nodes for different input dimensions under gradient flow with finite data. The shaded region indicates the standard error of the binomial proportion. (\textbf{C}) Angle at convergence between the student and teacher nodes for unsuccessful seeds. The shaded region indicates standard error of mean. (\textbf{D}) Same as in (A), but for a two-neuron bias-free teacher-student system and showing the presence of local (\textit{red dots}) and global minima (\textit{black dots}). The two teacher nodes are orthogonal, with opposite readout weights. Student trajectories are sampled from a grid of possible initial angles, measured with respect to the first teacher node. (\textbf{E}) Fraction of the loss landscape in (D) leading to a global minimum evaluated for 100 starting trajectories as a function of the angle $\theta^\star$ between the two teacher neurons. The setting from (D) corresponds to $\theta^\star=\frac{1}{2} \pi$. Remarkably, there are no local minima when both teacher nodes have the same sign. (\textbf{F}) Fraction of students reaching the global minimum as a function of the angle between the two teacher nodes for gradient flow with finite data. The shaded region indicates the standard error of the binomial proportion.}
    \label{fig:ana-combined}
\end{figure}

Under these assumptions, the loss landscape has no spurious local minima: all trajectories converge to the global minimum. In particular, there exist no OOB minima, because infinite Gaussian data is unbounded. However, if the starting angle between the teacher and student neuron is high, the student bias takes a long trajectory out before convergence (see Figure~\ref{fig:ana-combined}A), indicating that it may go out of bounds in realistic cases where data is bounded. 
Indeed, if we remove the assumptions of fast readout weights and infinite data and simulate the single neuron under standard gradient flow, success rates depend strongly on starting similarity (see Figure~\ref{fig:ana-combined}B). Neurons that start too dissimilarly are rotated away from the teacher neuron (Figure~\ref{fig:ana-combined}C) and driven out of bounds (Figure~\ref{sfig:single-neuron-kink-diverged}A); furthermore, no neurons that start with the wrong readout sign successfully match the teacher neuron (Figure~\ref{sfig:single-neuron}B).

\subsection{Two-neuron teacher-student setting}

The two-neuron teacher-student setting has $2D+5$ parameters, composed of $2D$ embedding weights, two bias terms, two readout weights, and an outer bias. To reduce the number of parameters, we again assume infinite data and fast readout weights. Further assuming bias-free teacher and student networks and ReLU nonlinearity, we obtain a loss landscape in terms of the two angles $\theta_1$ and $\theta_2$ between the student weights and the first teacher weight (see Appendix~\ref{app:analytical-reduction} for full derivation),
\begin{equation}
    \mathcal{L}(\theta_1,\theta_2) = \frac{1}{2} \vec a^{\star \T} \Big[ \vec{K}_{TT} - \vec{K}_{TS} \, \vec{K}_{SS}^{-1} \, \vec{K}_{TS}^\T \Big] \vec a^\star,
    \label{eq:schur-loss}
\end{equation}
where $\vec{K}_{TT}=\langle \sigma(\vec z^\star) \sigma(\vec z^\star)^\T \rangle_{\vec z^\star}$, $\vec{K}_{SS}=\langle \sigma(\vec z) \sigma(\vec z)^\T \rangle_{\vec z}$, $\vec{K}_{TS}=\langle \sigma(\vec z^\star) \sigma(\vec z)^\T \rangle_{\vec z^\star, \vec z}$ are the teacher--teacher, student--student, and teacher--student kernel matrices. Due to the Gaussian data, $\vec z,\vec z^\star$ are jointly Gaussian with zero mean, unit variance, and covariances defined by the angles between the respective embedding weights; using the rotational invariance of the setup we set the first teacher as the reference such that all angles can be expressed via $\theta_1,\theta_2$, and the angle between the two teacher weights $\theta^\star$. The teacher's readout weights $\vec a^{\star}$ and the angle of their weights $\theta^\star$ jointly control the geometry of the loss landscape.

For \emph{same-sign readouts} ($\vec a^\star = (1, 1)^\T$, corresponding to teachers in the minimally dissimilar regime) and fast readout weights, the loss landscape has no local minima other than the global minimum: all gradient flow trajectories converge regardless of initialization (see Figure~\ref{sfig:loss-landscape-2node}). In contrast, for \emph{mixed-sign readouts} ($\vec a^\star = (-1, 1)^\T$, corresponding to teachers in the maximally dissimilar regime), the landscape develops local minima (Figure~\ref{fig:ana-combined}D). At the local minima, both student neurons align to the same teacher neuron, leaving the other unmatched. The basin of attraction of these local minima decreases with increasing $\theta^\star$ (Figure~\ref{fig:ana-combined}E).

We again juxtapose this analysis of the loss landscape for fast readout weights and infinite data with the empirical setting with finite data, Glorot-initialized weights, and standard gradient flow (Figure~\ref{fig:ana-combined}F). We see a similar pattern here for teachers with opposite readout signs: As the angle between the teacher neurons $\theta^\star$ grows, a larger fraction of students get stuck in local minima above the global minimum, with the exception of perfectly parallel hyperplanes when $\theta^\star=0$. 
For teachers with the same signs the result is strikingly different to the analytical setting: Instead of the single global minimum in the analytical setting, there is a significant number of local minima such that the overall success rate is slightly smaller than in the case of opposing signs.

The changing structure of the loss landscapes therefore provides a first explanation for why the local minima are more prominent in the case of dissimilar teacher nodes. The difference in success rates between the analytical and empirical setting motivates the hypothesis that teacher nodes with the same readout sign increase the importance of OOB minima, which are absent in the analytical setting.
Finally, the high success rates achieved by students with fast readout weights motivate the use of differential learning rates and different optimizers, which we explore in Section~\ref{sec:differential-learning-rates}. 

\section{Analysis of the local minima}

\label{sec:minima}

\begin{figure}
    \includegraphics[width=\textwidth]{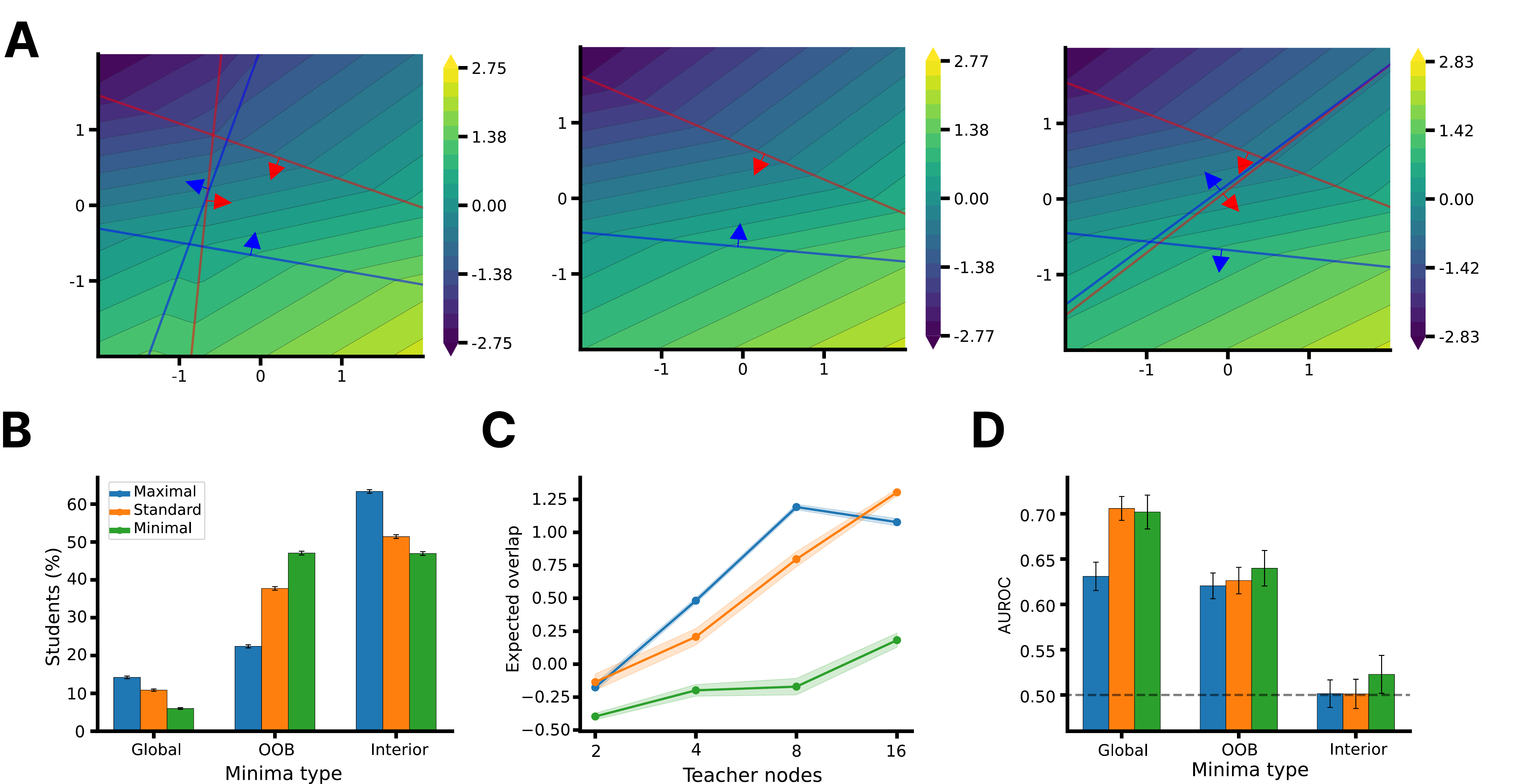}
    \caption{\textbf{Analyzing the local minima reached by the system for $\rho=1$.} (\textbf{A}) Example solutions reached by neurons in (\textit{left}) the global minimum, (\textit{middle}) a local minimum with two out-of-bounds (OOB) neurons pointing inwards (beyond the domain visible in the figure), (\textit{right}) an interior local minimum in which all nodes are in bounds ($\rho=1, D=2, M=4$). (\textbf{B}) The frequency of types of minima -- global, OOB, and interior -- across teacher distributions. The percentage across the three different types of minima are computed within each setting for $M$ and $D$, and then aggregated across settings. Error bars represent standard error of mean across the eight settings $M \in \{2,4,8,16\}$ with $D=2$ fixed and $D \in \{2, 4, 8, 16\}$ with $M=4$ fixed. (\textbf{C}) Initial overlap between teacher and student as a function of the teacher distribution and number of teacher nodes. (\textbf{D}) AUROC scores for 1-v-all classification of the minima type based on the initial overlap as the decision score for a single teacher, and aggregated across all teachers and settings. Error bars represent standard error of mean across different teachers.}
    \label{fig:minima}
\end{figure}

Going back to networks with more than two hidden neurons, we confirm that global minima, out-of-bounds (OOB) local minima, and interior local minima are found from different student initializations (Figure~\ref{fig:minima}A). Confirming the hypothesis we formed in the previous section, students trained on teachers with the maximally dissimilar distribution (which reach the global minimum the most frequently) have fewer OOB neurons than students trained on teachers of the other types, while having a larger number of students in interior local minima (see Figure~\ref{fig:minima}B for average across all settings and  Figure~\ref{sfig:minima-types} for individual runs). These observations are qualitatively consistent with the analysis of Section~\ref{sec:loss-landscapes}.%

To evaluate whether these differences are the result of the differences in readout sign or of orientation of the teacher nodes, we repeated the experiment for two additional sets of distributions, maximally dissimilar teachers with same readout signs, and minimally dissimilar teachers with evenly-matched readout signs across settings. Changing the signs had a large impact on success rates, yet neither was as successful as the maximally dissimilar distribution, indicating that both diverse orientations and diverse readout signs are necessary to drive high success rates (Figure \ref{sfig:signs-sweep}). Both of these additional distributions had success rates only slightly above that of the teachers with minimally dissimilar nodes. However, their failure modes were very different. Teachers with same-signed nodes consistently led to a high rate of OOB neurons, while teachers with mixed nodes consistently led to more interior minima (Figure~\ref{sfig:signs-minima}).

Next, we study the impact of introducing symmetries. The hyperplanes of the teachers we study all have small offsets that prevent symmetries between the different hyperplanes. When hyperplanes are perfectly parallel, however, students sometimes have additional degrees of freedom that they can exploit in matching teachers. For instance, two teacher nodes that are parallel but have opposite readout signs can be matched by student nodes from a multitude of different configurations, as long as the student nodes are also parallel and have opposite readout signs. This should increase the success rate of students. To investigate the effect this would have, we trained another set of 10 teachers with perfectly parallel hyperplanes, i.e. the minimally dissimilar distribution with $\theta=0$, on the $D=2, M=4$ setting. We find that the success rate for this distribution is indeed higher than for the non-parallel case. Moreover, the reduction comes specifically from a decrease in interior minima, with student neurons going out of bounds at a higher rate (Figure~\ref{sfig:parallel}). Compared with the minimally dissimilar distribution, therefore, the symmetries reduce the number of interior minima, while still pushing students that are initialized incorrectly out of bounds.

To study the effect of initialization on success, we computed the overlap term $\max_{p \in P_N} \sum_{j=1}^M a_{p(j)} a_j^\star \, \vec{w}_{p(j)}^\T \vec{w}_j^\star$ from Equation~\ref{eqn:initial-overlap} for each individual teacher-student pair. This initial overlap is highest for the maximally dissimilar teacher distribution, and tends to increase with increasing number of teacher nodes (Figure~\ref{fig:minima}C), and roughly trends towards zero for increasing dimensionality (Figure~\ref{sfig:overlap-dimensionality}). We hypothesized that a higher initial overlap would lead to a higher success rate in finding the global minimum (see Figure~\ref{sfig:auroc}A for a scatter plot showing the relationship for a single seed). To test this hypothesis, we used the initial overlap as the decision score in an AUROC calculation to see how well we could use it to classify what minimum type that teacher reaches (in a 1-v-all manner). We computed a separate AUROC score for each teacher across the 50 trained students. We averaged the results first within an $M, D$ setting (see Figure~\ref{sfig:auroc}), and then across settings (see Figure~\ref{fig:minima}D). Consistently across settings, the initial similarity predicts success rate better than chance for global and OOB minima. However, it performs only at around chance level for interior minima, indicating a complicated relationship in this latter type.

\section{Differential learning rates}

\label{sec:differential-learning-rates}

\begin{figure}
    \centering    \includegraphics[width=\textwidth]{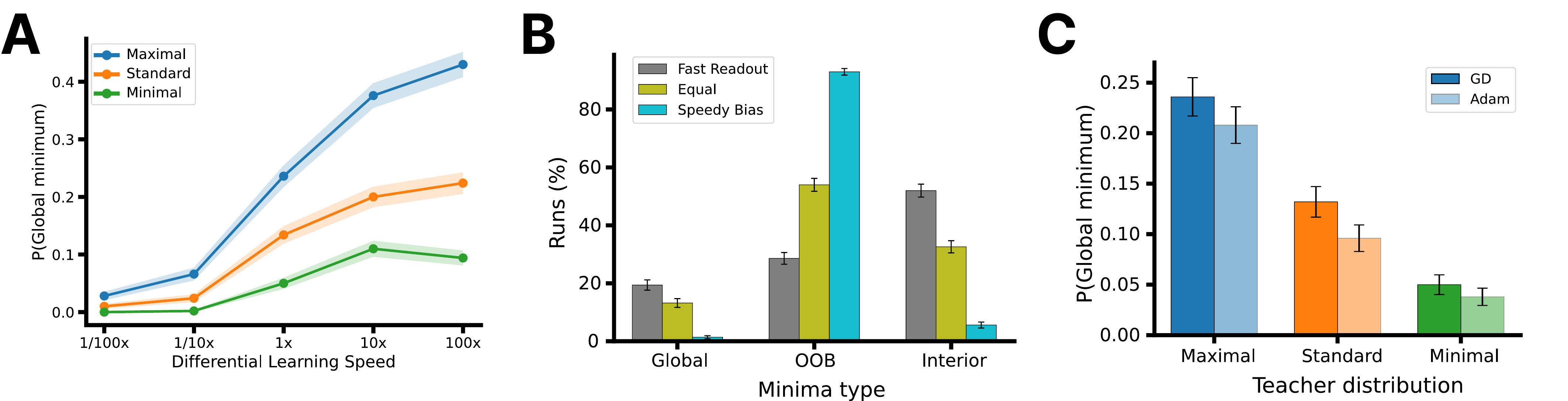}
    \caption{\textbf{Differential learning rates for $\rho=1$.} (\textbf{A}) Percentage of student networks that reach the global minimum as we differentially change, in opposite directions, the speed of the hidden layer bias and of the readout weights and bias. A differential learning speed of 10x corresponds to increasing the speed of the readout weight and bias by a factor of 10, and decreasing the speed of the inner bias by a factor of 10. The shaded region indicates the standard error of the binomial proportion. (\textbf{B}) Frequency of the different types of minima -- global, out of bounds, and interior -- for (\textit{grey}) fast readout weights, (\textit{light green}) equal learning rates, and (\textit{light blue}) fast biases in the $M=4$, $D=2$ setting. Error bars represent standard error of mean across the 10 different teachers. (\textbf{C}) Fraction of networks finding the global minimum under (\textit{dark blue}) gradient flow and (\textit{light blue}) the Adam optimizer. The error bars indicate the standard error of the binomial proportion.}
    \label{fig:tauinv}
\end{figure}

The loss landscape analyses have shown that local minima disappear under fast readout weights; the empirical simulations have shown that neurons go frequently out of bounds for incorrect starting positions. These two observations suggest that simulation performance could potentially improve with differential learning speeds, in which the readout weights are learned more quickly than the embedding weights and the inner bias. %
We hypothesized that success rates could be increased by applying different learning rates to different parameters, so as to align the readout sign before the neurons go out of bounds. We create versions of the optimization with different learning rates for different parameters. When we increase the learning rate of the readout weights for sample networks, success rates increase. In contrast, success rates are decreased when the learning rate of the inner bias $b$ is increased, as more neurons are driven out of bounds (Figure~\ref{fig:tauinv}A). These changes are driven by corresponding de- and increases in the number of OOB neurons (Figure~\ref{fig:tauinv}B). In contrast, changing the optimizer from gradient flow to Adam does not increase the success rates (Figure~\ref{fig:tauinv}C). This indicates that the momentum effect plays other roles and may help mitigate the impact of other critical points or noise in stochastic gradient descent, but does not prevent neurons from going out of bounds.

\section{Discussion}

In this paper, we studied teacher-student learnability --- the probability that gradient-based optimization finds the global minimum --- as a function of the teacher geometry, i.e. the configuration of weight vectors in the teacher network. 

The maximally dissimilar and minimally dissimilar distributions correspond to two different hypotheses of what makes a teacher difficult. The minimally dissimilar distribution could be considered ``easy" because the different teachers have a strong shared component, which could drive student neurons to quickly match this primary direction followed by gradual differentiation to the individual teacher nodes. The maximally dissimilar distribution could be considered ``easy" because the different teacher nodes are as diverse as possible and thus can correspond to different (randomly-initialized) student neurons. Our results fall strongly on the side of the latter hypothesis: Networks benefit strongly from having signals that are disentangled. For ReLU networks, this is maximized for weight vectors pointing in opposite directions.

Our results indicate a complicated interplay of teacher geometry and student initialization.
Starting similarity, which is correlated with success within an individual teacher, is higher for maximally dissimilar teacher nodes, as randomly initialized student nodes tend to be diverse (see Appendix~\ref{app:res-proof}).
Using the popular notion of lottery tickets in the context of neural networks \cite{frankle.carbin2018,saraomannelli.etal2020,tian.etal2019}, and interpreting tickets as a set of units (instead of a subgraph), one could see the maximally dissimilar teacher distribution together with standard student initialization as a lottery with more winning tickets than the other distributions.
However, the expected initial teacher-student overlap also does not fully explain the rate of finding the global minimum. Convergence to interior local minima seems impossible to predict based on starting similarity (Figure~\ref{fig:minima}).
Probably, effects on the dynamics also have an influence, like decoupling the student neurons' trajectories from each other by pulling them in different directions, leading to a ``neural race" in which each student neuron converges to the closest target feature independently~\cite{boursier2022gradient,boursier.flammarion2025,pinson2026s}. We expect this decoupling to occur primarily for students trained on maximally dissimilar teachers as they are pulled in different directions. This perhaps is one of the major factors explaining the difference between the maximally and minimally dissimilar teachers.

Surprisingly, we find that the rate of OOB neurons can be decreased substantially by increasing the speed of the readout weights, and slowing down the speed of the biases. This simple intervention increases success rates substantially. In the literature, differential learning rates across layers are frequently encountered in the domain of transfer learning, with the intuition that early layers will already have learned useful features during pretraining, so that readout weights are given a higher learning rate~\cite{yosinski.etal2014, howard.ruder2018, zhang.etal2020, yang.etal2019a, clark.etal2019}. However, only few works have considered applying this same intervention already during initial feature learning~\cite{marion.berthier2023, zeng.etal2026, galashov.etal2026}. In the teacher-student setting we consider, we see how allowing the bias term to adapt too quickly in difficult learning settings hinders the formation of good features, in particular when there is an initial mismatch in the readout sign; the solution is slowing down the bias term while increasing the speed of the readout weights. Applying this to real-world domains represents a promising direction for future research.

Overall, this study provides a framework for understanding why some functions are easier to learn for neural networks than others, and to being able to train smaller networks by limiting rates of network failure.

\begin{ack}
We thank Yatin Dandi for thoughtful discussions. This work was supported by SNF Project 200021-236436. 
\end{ack}

\clearpage

\printbibliography

\newpage

\appendix

\section{Related works}

\label{app:related-works}

\subsection{Characterizing fixed points in loss landscapes for teacher-student systems}
Fukumizu and Amari described the formation of lines of critical points that emerge from adding neurons to a smaller network at a critical point of its loss ~\cite{fukumizu.amari2000}. Simsek et al.~\cite{simsek.etal2021} built on this in the teacher-student setting, showing that permutation symmetries generate a precisely characterised hierarchy of symmetry-induced saddles, so that the global minima manifold grows faster than the saddle manifold as width increases, making overparameterization a systematic mechanism for landscape smoothing. Interestingly, the ratio of all symmetry-induced critical subspaces to global minimum subspaces drops below one when the student network has approximately 1.5 times the number of hidden neurons as the teacher network, independently of the distribution from which the teacher weights are drawn. Wu et al. sharpened this picture for shallow ReLU-like networks, introducing the notion of ``escape neurons" that satisfy specific first-order conditions and allow overparameterized networks to escape from saddle points~\cite{wu.etal2024}. An important factor influencing convergence rates is the number of permutation-induced symmetries, which are particularly prevalent for ReLU-activation functions~\cite{martinelli.etal2024}. In this paper, we build on these studies to investigate how these factors are affected by differences between different teachers. However, we don't primarily focus on symmetries, considering teacher distributions without parallel hyperplanes and students with small amounts of overparameterization.

\subsection{Learnability of functions}

Most prior work on learnability focuses on how quickly neural networks can learn to approximate a given function~\cite{zhang.etal2017}. The computational cost of learning has been shown to depend on the kind of function. The index of the first non-zero Hermite coefficient of the target function, known as the information exponent, governs sample complexity~\cite{arous.etal2021}. A class of quickly learnable functions for two-layer networks are those that satisfy the so-called staircase~\cite{abbe.etal2021} or merged-staircase~\cite{abbe.etal2022} properties, whose higher-order Fourier functions cannot be built from lower-order ones requiring superpolynomially more updates. 

In this paper, however, we are primarily interested in the limit of how well a function will be implemented at convergence, and how many neurons this requires. For this reason, we use a two-step optimization procedure to ensure that the student networks converge to the local minima whose basin of attraction they fall under. The most classic finding of this category is that XOR cannot be implemented by a single perceptron~\cite{minsky.papert1969}, but instead requires a larger network~\cite{rumelhart.etal1986}. Its generalization to higher dimensions, the parity function, is still considered the classic example of a difficult function today~\cite{daniely.malach2020, shoshani.shamir2025, barak.etal2022}, as are periodic functions~\cite{ziyin.etal2020}. We consider a more abstract setting, but periodic functions have some similarity to our minimally dissimilar nodes.

\subsection{The ReLU activation function}

The ReLU activation function, on which this paper focuses, partitions input space into linear regions~\cite{montufar.etal2014, pascanu.etal2013, serra.etal2018}. Tian~\cite{tian2017} exploits this piecewise linearity to derive a population gradient characterizing the evolution of ReLU networks. In later work, they show that student neurons initialized close to teachers converge faster~\cite{tian.etal2019}. We in part build on these results in formalizing the different teacher distributions, but we focus on the implications of this finding for the learnability of different kinds of teachers. Lee et al~\cite{lee.etal2024b} have analyzed the dynamics of single-neuron ReLU networks in detail. In this paper, we take a similar view, but focus on studying the loss landscape.

\clearpage
\section{Training details}

\subsection{Training}

\label{app:training}

We train neural networks by solving the gradient flow equation with the MLPGradientFlow package~\cite{brea.etal2023}. For each teacher, we fit 50 different student seeds. The input consists of 10,000 points chosen from a standard multivariate Gaussian distribution of dimension $D$.

Training proceeds in two phases: First, an ODE solver is run until convergence, for up to two hours of CPU time (see Appendix~\ref{app:compute}); second, a second-order optimizer optimizes the parameters for an additional two hours.

Some runs, in particular in the larger settings, did not terminate successfully within the timeframe. When no file was generated, the runs were excluded from analysis; Table~\ref{stab:convergence-summary} reports the number of completed runs for each setting. All runs that were successfully saved were included in the analysis without application of an additional convergence criterion.

\subsection{Weight normalization procedure}

For the maximally and minimally dissimilar distributions, each embedding weight vector $\vec{w}_j^\star$ is constructed to lie on the unit hypersphere by design. For the standard distribution, weight vectors are drawn with scale $1/\sqrt{D}$, giving expected unit norm.

In the Python analysis pipeline, teacher networks are additionally normalized for visualization and overlap computation: each embedding weight vector is rescaled to unit norm and multiplied by 3. The readout weights are then iteratively adjusted so that the standard deviation of the teacher output on a 30,000-sample standard Gaussian input equals 1 and the mean equals 0. This normalization is applied only in the analysis scripts and does not affect the training experiments.

\subsection{Convergence criteria and success classification}

A student run is classified as having reached the global minimum if its final training loss satisfies $\mathcal{L} < 10^{-18}$. Training itself terminates when the loss reaches $10^{-30}$, when a patience criterion is met (no improvement over $10^6$ iterations), or when the time limit is reached (2 hours for ODE-based training, 2 hours for second-order optimization).

For student runs that do not reach $\mathcal{L} < 10^{-30}$, the minimum eigenvalue of the Hessian is computed at the stopping point. Runs with minimum eigenvalue $> 10^{-5}$ are classified as having converged to a strict local minimum (\texttt{:strict}); all others are classified as non-strict critical points or saddle regions (\texttt{:nonstrict}). Runs where the optimizer's line search failed are recorded separately (\texttt{:opt\_fail}).

\subsection{Initial overlap computation}
\label{app:meth-overlap}

The overlap between a student neuron $i$ and a teacher neuron $j$ is computed as
\begin{equation}
    s_{ij} = a_i \, a_j^\star \, \left( \vec{w}_i^\T \vec{w}_j^\star \right),
    \label{eq:pairwise-overlap}
\end{equation}
the product of the two readout weights and the inner product of the embedding weight vectors. A higher score indicates that the student neuron has a similar weight direction and the same readout sign as the teacher neuron. Note that a double-negative case (both $a_i \, a_j^\star < 0$ and $\vec{w}_i^\T \vec{w}_j^\star < 0$) is treated as no overlap for ReLU networks, since the two sign flips do not cancel functionally.

The optimal assignment between student and teacher neurons is found using the Hungarian algorithm, maximizing the total overlap score $\sum_{j=1}^M s_{p(j),j}$ over all permutations $p \in P_N$. This is exactly the quantity $\max_{p \in P_N} \sum_{j=1}^M s_{p(j),j}$ whose expectation over $\mathcal{T}$ and $\mathcal{I}$ defines $S(\mathcal{T}, \mathcal{I})$ in Equation~\ref{eqn:initial-overlap}.

\subsection{Compute used}
\label{app:compute}

The experiments were run on a CPU across 20 parallel cores. The paper requires about 100,000 individual random seeds. Each seed was capped at 4 hours, though most finished much faster, giving an estimated total runtime of about 2000 hours (on AMD EPYC 9454 48-Core Processor CPU) with this setup.

\section{Teacher distributions}

\subsection{Teacher generation details}

\label{app:teacher-construction}

We sample the teachers from the distributions with a fixed random seed (0). We generate 10 teachers per combination of input dimensionality $D \in \{2, 4, 8, 16\}$ and number of teacher nodes $M \in \{2, 4, 8, 16\}$.

Throughout this subsection $j = 1, \dots, M$ indexes the teacher nodes, so that $\vec{w}_j^\star$, $b_j^\star$ and $a_j^\star$ denote the embedding weight vector, the inner bias, and the readout weight of teacher node $j$, and $c^\star$ denotes the teacher outer bias.

\textbf{Maximally dissimilar.} The weight vectors $\vec{w}_j^\star$ are initialized on a quasi-uniform hyperspiral to be approximately equidistant on the unit hypersphere. They are then refined by $10^6$ gradient descent steps with learning rate 0.1 minimizing the sum of squared pairwise cosine similarities, with renormalization to the unit sphere after each step. Gaussian noise with standard deviation 0.1 is added and the vectors are renormalized. The biases $b_j^\star$ are sampled independently and uniformly from the discrete set $\{-2/3,\,-1/3,\,0,\,1/3,\,2/3\}$, with the constraint that two approximately anti-parallel nodes (cosine similarity $< -0.99$) may not have biases that sum to zero, to avoid hidden overparameterization symmetries. The readout weights $a_j^\star$ are set to an equal number of $+1$s and $-1$s, randomly shuffled; the outer bias is $c^\star = 0$.

\textbf{Standard.} The entries of the weight vectors are sampled as $w_{jd}^\star \sim \mathcal{N}(0, 1/D)$ independently for each node $j$ and dimension $d$, i.e.\ with standard deviation $1/\sqrt{D}$. All biases are zero, $b_j^\star = 0$. The readout weights are sampled as $a_j^\star \sim \mathcal{N}(0, 1)$ independently; the outer bias is $c^\star = 0$.

\textbf{Minimally dissimilar.} A base direction $\bar{\vec{w}}$ is drawn uniformly from the unit hypersphere. The weight vectors $\vec{w}_j^\star$ are then drawn uniformly at random and accepted only if the angle $\theta$ to $\bar{\vec{w}}$ satisfies $\theta < \pi/8$, repeating until $M$ vectors are accepted. The biases $b_j^\star$ are set to $M$ evenly spaced values in $[-0.5, 0.5]$ and assigned to nodes in random order. All readout weights are $a_j^\star = +1$ and the outer bias is $c^\star = 0$.

\subsection{The different teacher distributions maximize and minimize expected initial similarity}

\label{app:res-proof}

In this section, we show that the initial similarity is maximized by maximally dissimilar teachers, and minimized by minimally dissimilar ones, in a simplified setting in which we start by considering only the orientation of the teacher and student nodes given by $\vec{w}^\star$ and $\vec{w}$.

Consider the expected similarity function for teachers $\vec{w}_1^\star, \vec{w}_2^\star \in \mathbb{S}^1$, where $\mathbb{S}^1$ denotes the unit circle,
\begin{equation}
    s(\vec{w}_1^\star, \vec{w}_2^\star) \;=\; \mathbb{E}_{\vec{w}_1, \vec{w}_2 \sim \mathrm{Unif}(\mathbb{S}^1)}\!\left[\max_{p \in P_2}\,\sum_{j=1}^{2} \vec{w}_{p(j)}^\T \vec{w}_j^\star\right],
    \qquad P_2 = \{\mathrm{id}, \mathrm{swap}\}.
\end{equation}

The two permutations give
\begin{align}
    S_\mathrm{id}   &\;=\; \vec{w}_1^\T \vec{w}_1^\star + \vec{w}_2^\T \vec{w}_2^\star, \\
    S_\mathrm{swap} &\;=\; \vec{w}_2^\T \vec{w}_1^\star + \vec{w}_1^\T \vec{w}_2^\star,
\end{align}
so that $s(\vec{w}_1^\star, \vec{w}_2^\star) = \mathbb{E}[\max(S_\mathrm{id}, S_\mathrm{swap})]$.
By rotational invariance of the student distribution, $s$ does not depend on the absolute positions of the teacher vectors but only on the angle $\theta$ between them. We therefore set $\vec{w}_1^\star = (1,0)^\T$ and $\vec{w}_2^\star = (\cos\theta, \sin\theta)^\T$ without loss of generality.

Applying $\max(a,b) = \tfrac{a+b}{2} + \tfrac{|a-b|}{2}$ and using
\begin{equation}
    \mathbb{E}[S_\mathrm{id} + S_\mathrm{swap}] \;=\; (\vec{w}_1^\star + \vec{w}_2^\star)^\T \, \mathbb{E}[\vec{w}_1 + \vec{w}_2] \;=\; 0,
\end{equation}
we obtain
\begin{equation}
    s(\theta) \;=\; \tfrac{1}{2}\,\mathbb{E}\bigl|S_\mathrm{id} - S_\mathrm{swap}\bigr|.
\end{equation}
Direct expansion yields
\begin{equation}
    S_\mathrm{id} - S_\mathrm{swap}
    \;=\; (1-\cos\theta)(w_{1,x} - w_{2,x}) - \sin\theta\,(w_{1,y} - w_{2,y})
    \;=\; \vec{v}^\T \vec{d},
\end{equation}
where $\vec{v} = (1 - \cos\theta,\, -\sin\theta)^\T$ and $\vec{d} = \vec{w}_1 - \vec{w}_2$, and $w_{i, x}$ and $w_{i, y}$ denote the $x$ and $y$ components of $\vec{w}_i$, respectively. Then
\begin{equation}
    \|\vec{v}\|^2 \;=\; (1-\cos\theta)^2 + \sin^2\theta \;=\; 2(1-\cos\theta) \;=\; 4\sin^2(\theta/2),
\end{equation}
so $\|\vec{v}\| = 2\,|\sin(\theta/2)|$. The distribution of $\vec{d}$ is rotationally invariant, so $\mathbb{E}|\vec{u}^\T \vec{d}|$ is independent of $\vec{u}$ for the unit vector $\vec{u} := \vec{v} / \|\vec{v}\|$. The expectation $\mathbb{E}|\vec{u}^\T \vec{d}|$ does not depend on $\theta$ since $\vec{d}$ is rotationally invariant. We therefore rotate $\vec{u}$ to align with the $x$-axis:
\begin{equation}
    \mathbb{E}|\vec{u}^\T \vec{d}| = \mathbb{E}|\vec{e}_1^\T \vec{d}| = \mathbb{E}|w_{1,x} - w_{2,x}| =: C > 0.
\end{equation}
Combining with $\|\vec{v}\| = 2|\sin(\theta/2)|$ gives
\begin{equation}
    s(\theta) = C\,|\sin(\theta/2)|.
\end{equation}

This expression is minimized at $\theta = 0$ and maximized at $\theta = \pi$. These two settings correspond to the orientations between nodes of the minimally dissimilar teacher distribution and maximally dissimilar teacher distribution, respectively. 

In the settings we consider in the paper, the readout signs also differ between the maximally dissimilar distribution of teachers (where they are evenly split between $\{+1, -1\}$) and the minimally dissimilar distribution (where every node has $a_j^\star = +1$). Randomly-initialized student weights will randomly be positive or negative. By a similar argument, the similarity at initialization will therefore also be greater with the maximally dissimilar teacher distribution. 

We describe this result only for a simplified setting in $D=2$, the starting similarity can be seen to differ empirically between the different distributions Figure~\ref{fig:minima}C and Figure~\ref{sfig:overlap-dimensionality}.

\section{Existence of different types of out-of-bounds minima}
\label{app:minima-types}

\label{app:boundary-dynamics}

We consider training a neural network with ReLU activation function on a finite dataset with a maximum distance of $R$ to the origin; that is, $X \in \{\vec{x} : \|\vec{x}\| \leq R\}$. Recall that for a neuron $\vec{w}$ the kink $k = -\frac{b}{||w||}$ is the signed distance from the origin to its hyperplane along its weight vector $\vec{w}_j$, and that the neuron is considered out-of-bounds if $k \leq \min_i \vec{w} \cdot \vec{x}_i$ or $k \geq \max_i \vec{w} \cdot \vec{x}_i$. If $k \geq \max_i \vec{w} \cdot \vec{x}_i$, the neuron is a \textit{dead neuron}; that is, it is not active for any data point. This is a local minimum because the neuron carries no gradient signal. 

In contrast, if $k \leq \min_i \vec{w} \cdot \vec{x}_i$, it is a \textit{linear neuron} that is active for every data point. Writing $\varepsilon_i = y_i - y_i^\star$ for the residual, $\partial \mathcal{L}/\partial b = a\,\langle \varepsilon \rangle$ and $\partial \mathcal{L}/\partial \vec{w} = a\,\langle \varepsilon\,\vec{x} \rangle$ since $\sigma'(\vec{w}^\T\vec{x}+b) = 1$ everywhere in the always-on regime. Differentiating $k = -b/\|\vec{w}\|$ along gradient flow yields
\begin{equation}
\dot{k} = \frac{\eta}{\|\vec{w}\|}\left( \frac{\partial \mathcal{L}}{\partial b} + k\,\frac{\partial \mathcal{L}}{\partial \|\vec{w}\|} \right),
\qquad
\frac{\partial \mathcal{L}}{\partial \|\vec{w}\|} = \frac{\vec{w}^\T}{\|\vec{w}\|}\,\frac{\partial \mathcal{L}}{\partial \vec{w}} = a\,\left\langle \varepsilon\,\frac{\vec{w}^\T\vec{x}}{\|\vec{w}\|} \right\rangle.
\end{equation}
The tangential component of $\dot{\vec{w}}$ rotates the neuron's orientation but drops out of $\dot{k}$, which depends only on $b$ and $\|\vec{w}\|$.

Setting $k = -R$, the neuron re-enters the data domain ($\dot{k} > 0$) iff
\begin{equation}
a\left( \langle \varepsilon \rangle - R\,\left\langle \varepsilon\,\frac{\vec{w}^\T\vec{x}}{\|\vec{w}\|} \right\rangle \right) > 0.
\label{eq:reentry}
\end{equation}
An error term $\langle \varepsilon \rangle$ and a weight-norm term $R\,\langle \varepsilon\,\vec{w}^\T\vec{x}/\|\vec{w}\|\rangle$ compete; the readout sign $a$ flips both. A neuron can therefore be trapped not because the residual lacks structure along $\vec{w}$, but because its readout has the wrong sign to exploit that structure, making this also a true local minimum. For large $R$ the bias term is negligible and re-entry is controlled by the sign of $-a\,\partial \mathcal{L}/\partial \|\vec{w}\|$ alone.

\section{Complete picture of the loss landscape of one- and two-node ReLU systems}

\subsection{Derivation}

\label{app:analytical-reduction}

Generally, it is very difficult to completely characterize the loss landscape of even small neural networks. Here, we achieve a full characterization of the loss landscape of specific ReLU students. To this end, we systematically reduce the number of parameters by making use of simplifying assumptions and symmetry arguments to just two variables. Thus, we are able to fully characterize the loss landscape of these systems as the teacher changes structure.

The assumptions we make to reduce the number of parameters are:

\begin{itemize}
    \item \textbf{Instantaneous readout weights}: the readout weights $\vec{a}$ evolve instantaneously. By assuming the embedding weights fixed, we can compute the optimal readout weights as their optimization is a convex problem. This eliminates a total of $N$ degrees of freedom from the student.
    \item \textbf{Homogeneity of the activation function}: due to the positive homogeneity of the ReLU activation ($\lambda\,\sigma(\vec{w}^\T \vec{x}) = \sigma(\lambda\,\vec{w}^\T \vec{x})$ for any $\lambda > 0$, so that $\|\vec{w}\|\,\sigma(\hat{\vec{w}}^\T \vec{x}) = \sigma(\vec{w}^\T \vec{x})$ with $\hat{\vec{w}} = \vec{w}/\|\vec{w}\|$) we can fix the norm of the student weight vectors to 1 without loss of generality, absorbing $\|\vec{w}\|$ into the readout weight. By parameterizing the student weights in terms of their angles, $N$ other degrees of freedom are eliminated.
    \item \textbf{Gaussian inputs} ($\vec{x} \sim \mathcal{N}(\vec{0}, \vec{I})$): the infinite support of the Gaussian distribution removes boundary effects that arise from finite data domains, and allows us to take the limit $K \to \infty$ so that the loss is computed exactly in expectation. This eliminates the case of out-of-bounds minima.
\end{itemize}

\paragraph{Gaussian data average.}
The starting point is the finite-sample square loss over $K$ data points,
\begin{equation}
    \mathcal{L} = \frac{1}{2K} \sum_{i=1}^{K} \big[ \vec{a}^\T \sigma(\vec{W} \vec{x}_i + \vec{b}) - \vec{a}^{\star \T} \sigma(\vec{W}^\star \vec{x}_i + \vec{b}^\star) \big]^2,
\end{equation}
where $\sigma(z) = \max(0,z)$ is the ReLU activation applied element-wise. For $\vec{x} \sim \mathcal{N}(\vec{0}, \vec{I})$, in the limit $K \to \infty$ the empirical average converges to an expectation over the input distribution:
\begin{equation}
    \mathcal{L} \;\to\; \frac{1}{2} \big\langle \big[ \vec{a}^\T \sigma(\vec{W} \vec{x} + \vec{b}) - \vec{a}^{\star \T} \sigma(\vec{W}^\star \vec{x} + \vec{b}^\star) \big]^2 \big\rangle_{\vec{x}}.
\end{equation}
Expanding the square and introducing the preactivations $\vec{z} = \vec{W}\vec{x}$ and $\vec{z}^\star = \vec{W}^\star \vec{x}$, which are jointly Gaussian with covariances $\langle \vec{z} \vec{z}^\T \rangle = \vec{W} \vec{W}^\T$, $\langle \vec{z} \vec{z}^{\star \T} \rangle = \vec{W} \vec{W}^{\star \T}$, and $\langle \vec{z}^\star \vec{z}^{\star \T} \rangle = \vec{W}^\star \vec{W}^{\star \T}$, the loss separates into
\begin{equation}
    \mathcal{L}(\vec{a}, \vec{W},\vec{b}) = \frac{1}{2} \vec{a}^\T \underbrace{\langle \sigma(\vec{z} + \vec{b}) \sigma(\vec{z} + \vec{b})^\T \rangle_{\vec{z}}}_{\vec{K}_{SS}} \vec{a} \;-\; \vec{a}^{\star \T} \underbrace{\langle \sigma(\vec{z}^\star + \vec{b}^\star) \sigma(\vec{z} + \vec{b})^\T \rangle_{\vec{z}, \vec{z}^\star}}_{\vec{K}_{TS}} \vec{a} \;+\; C,
\end{equation}
where $C = \frac{1}{2} \vec{a}^{\star \T} \vec{K}_{TT}\, \vec{a}^\star$ depends only on the teacher. The key observation is that each entry of these kernel matrices $[\vec{K}]_{ij} = \langle \sigma(\vec{w}_i^\T \vec{x} + b_i)\, \sigma(\vec{w}_j^\T \vec{x} + b_j) \rangle_{\vec{x}}$ depends on $\vec{w}_i$ and $\vec{w}_j$ only through their norms and their relative angle. For ReLU without bias, this expectation has the closed form~\cite{cho2009kernel}
\begin{equation}
    \langle \sigma(z_i) \sigma(z_j) \rangle = \frac{\|\vec{w}_i\| \|\vec{w}_j\|}{2\pi} \left[ \sin\theta + (\pi - \theta) \cos\theta \right], \qquad \cos\theta = \frac{\vec{w}_i^\T \vec{w}_j}{\|\vec{w}_i\| \|\vec{w}_j\|},
\end{equation}
which for unit-norm weights simplifies to $\frac{1}{2}f(r) = \frac{1}{2\pi}\!\left[\sqrt{1 - r^2} + (\pi - \cos^{-1}\!r)\,r\right]$ with $r = \cos\theta$. For ReLU with bias, the expectation can also be determined in closed form \cite{martinelli2025flat}.

\paragraph{Fast readout weights.}
Assuming the readout weights $\vec{a}$ evolve infinitely faster than the hidden-layer weights, they instantaneously converge to the optimum
\begin{equation}
    \vec{a}(\vec{W}, \vec{b}) = \langle \sigma(\vec{z} + \vec{b}) \sigma(\vec{z} + \vec{b})^\T \rangle_{\vec{z}}^{-1} \, \langle \sigma(\vec{z} + \vec{b}) \sigma(\vec{z}^\star + \vec{b}^\star)^\T \rangle_{\vec{z}, \vec{z}^\star} \, \vec{a}^\star
    \;=\; \vec{K}_{SS}^{-1} \, \vec{K}_{TS}^\T \, \vec{a}^\star .
\end{equation}
Substituting back, the loss reduces to a Schur complement:
\begin{equation}
    \mathcal{L}(\vec{W},\vec{b}) = \frac{1}{2} \vec{a}^{\star \T} \Big[ \vec{K}_{TT} - \vec{K}_{TS} \, \vec{K}_{SS}^{-1} \, \vec{K}_{TS}^\T \Big] \vec{a}^\star,
    \label{eq:schur-loss-app}
\end{equation}
where $\vec{K}_{TT}$, $\vec{K}_{SS}$, $\vec{K}_{TS}$ are the teacher--teacher, student--student, and teacher--student kernel matrices, respectively. For ReLU, the student weight norms cancel due to homogeneity, so the loss depends \emph{only on angles and biases}.

\paragraph{Single neuron with bias and $D>1$.}

With a single teacher and a single student neuron, the only angle in the problem is the angle $\theta$ between teacher and student weight. Thus, there are two degrees of freedom: the teacher-student angle $\theta$ and the student bias $b$.

The kernel matrices simplify to scalars, such that the loss reduces to
\begin{equation}
    \mathcal{L}(\theta,b) = \frac{1}{2} (a^{\star})^2 \Big[ \langle  \sigma(z^\star + b^\star)^2 \rangle_{z^\star} - \frac{\langle  \sigma(z + b)\sigma(z^\star + b^\star) \rangle_{z,z^\star}^2}{\langle  \sigma(z + b)^2 \rangle_{z}} \Big].
\end{equation}
Using homogeneity and introducing the kink $k=-b/\|\vec{w}\|$, we get
\begin{equation}
    \frac{\mathcal{L}(\theta,k)}{\frac{1}{2} (a^{\star})^2 \langle  \sigma(z^\star + b^\star)^2 \rangle_{z^\star}} =  1 - \frac{\langle  \sigma(z - k)\sigma(z^\star - k^\star) \rangle_{z,z^\star}^2}{\langle  \sigma(z - k)^2 \rangle_{z} \langle  \sigma(z^\star - k^\star)^2 \rangle_{z^\star}}
\end{equation}
where $z,z^\star$ are zero-mean and unit-variance with correlation $\langle z z^\star\rangle=\cos\theta$, and the denominator on the right hand side is a teacher-specific constant.

\paragraph{Two neurons without bias and $D=2$.}

In the two-neuron case, both teacher and student biases are set to zero for the analytical derivation of the loss landscape.

With $D=2$ input dimensions, a unit-norm weight vector in $\mathbb{R}^2$ is fully specified by a single angle on the unit circle. By rotational invariance of the Gaussian input distribution, we can fix the first teacher neuron at angle~$0$ without loss of generality. The angle $\theta^\star$ of the second teacher neuron is then a fixed parameter of the problem. The two remaining degrees of freedom are $\theta_1$ and $\theta_2$: the angles of the two student weight vectors relative to teacher neuron~1. These are the two axes of the landscape plots that follow. The loss~\eqref{eq:schur-loss-app} becomes
\begin{equation}
    \mathcal{L}(\theta_1, \theta_2) = \frac{1}{4} \vec{a}^{\star \T} \left[
    \begin{pmatrix} 1 & f(\cos\theta^\star) \\ f(\cos\theta^\star) & 1 \end{pmatrix}
    - \vec{T}_S^\T \, \vec{G}(\theta_1 - \theta_2) \, \vec{T}_S
    \right] \vec{a}^\star,
    \label{eq:loss-2d}
\end{equation}
where
\begin{equation}
    \vec{T}_S = \begin{pmatrix}
        f(\cos\theta_1) & f(\cos(\theta_1 - \theta^\star)) \\
        f(\cos\theta_2) & f(\cos(\theta_2 - \theta^\star))
    \end{pmatrix}, \qquad
    \vec{G}(\theta) = \frac{1}{1 - f(\cos\theta)^2}
    \begin{pmatrix} 1 & -f(\cos\theta) \\ -f(\cos\theta) & 1 \end{pmatrix}.
\end{equation}
In terms of the kernel matrices of Equation~\eqref{eq:schur-loss-app}, the leading matrix is $2 \vec{K}_{TT}$, while $\vec{T}_S = 2 \vec{K}_{TS}^\T$ and $\vec{G}(\theta_1-\theta_2) = \tfrac{1}{2} \vec{K}_{SS}^{-1}$; the matrix $\vec{G}$ is written with a separate symbol to avoid confusion with the teacher width $M$. This is a closed-form expression over the two-dimensional domain $(\theta_1, \theta_2) \in [-\pi, \pi]^2$, which can be fully visualized as a contour plot for any given $\vec{a}^\star$ and $\theta^\star$.

In the gradient flow simulations, we use a timescale hierarchy $\tau_a \ll \tau_W \ll \tau_b$: readout weights converge fastest, embedding weight angles evolve at an intermediate rate, and biases evolve slowest (initialized at zero). This means the trajectories first explore the zero-bias landscape rapidly via the angles, and only then slowly adjust the biases. If the zero-bias landscape contains local minima, the trajectories will reach them on the fast timescale; the slow biases then test whether these minima persist when the bias constraint is relaxed. The question becomes: does the landscape $\mathcal{L}(\theta_1, \theta_2)$ contain local minima, and if so, under what conditions on the teacher?

\subsection{Results}

\begin{figure}
    \centering
    \includegraphics[width=0.5\textwidth]{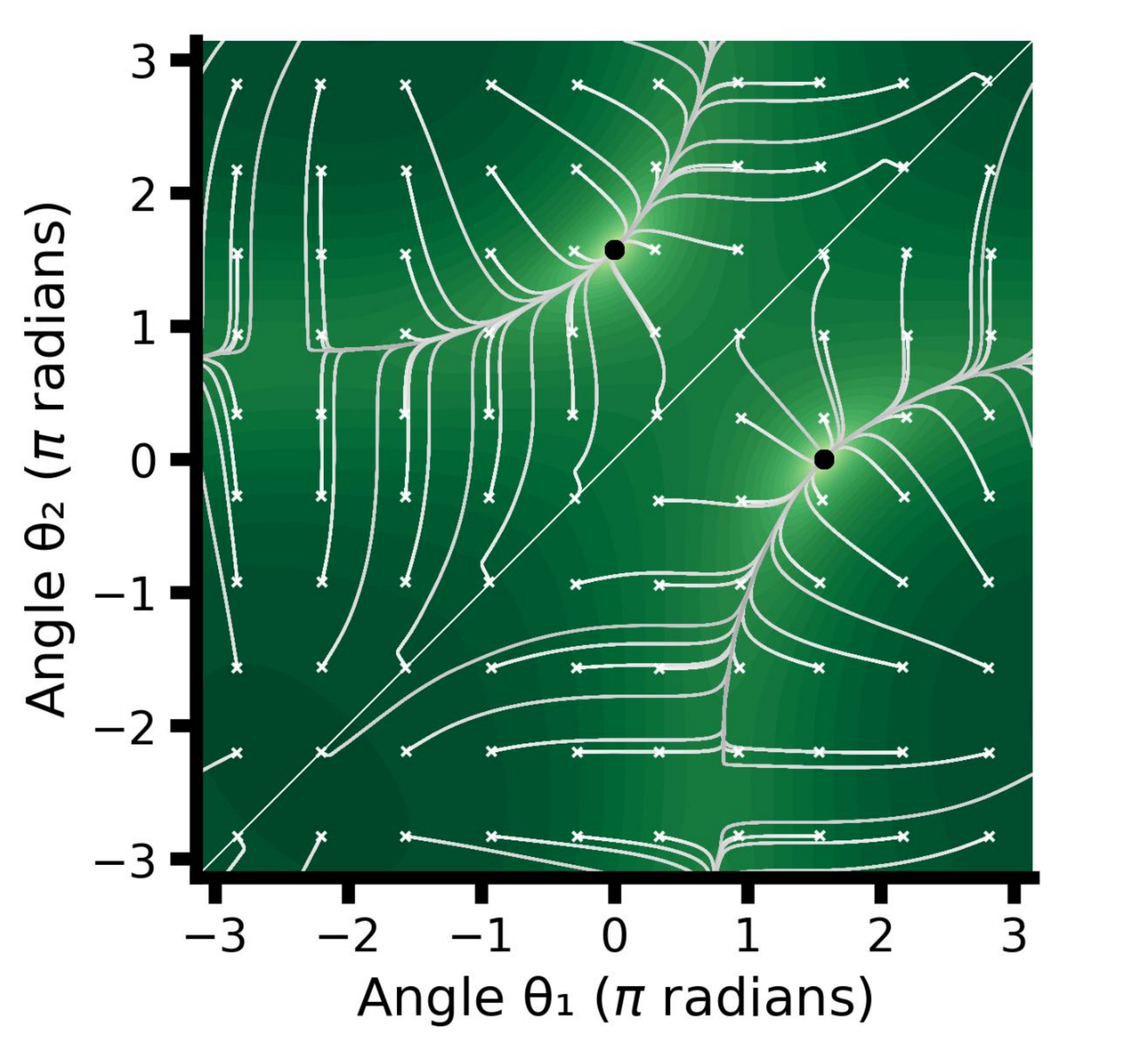}
    \caption{\textbf{Loss landscape and teacher output for the two-node ReLU system with $\theta^\star = \pi/2$ for same-sign readouts $\vec{a}^\star = (1,1)^\T$.} Zero-bias loss landscape~\eqref{eq:loss-2d} as a function of student angles $\theta_1$ and $\theta_2$ (log-scale, lighter = lower loss), with gradient flow trajectories overlaid in white. Students are initialized on a $10 \times 10$ grid covering all possible angle configurations (white crosses); trajectories converge to minima marked by black dots. All trajectories converge to a single global minimum (modulo permutation symmetry).}
    \label{sfig:loss-landscape-2node}
\end{figure}

We evaluate the loss landscape~\eqref{eq:loss-2d} for $\theta^\star = \pi/2$ under two teacher readout configurations and overlay gradient flow trajectories from a grid of random initializations (see Figure~\ref{sfig:loss-landscape-2node} for the same-sign readouts and Figure~\ref{fig:ana-combined}D for the mixed-sign readouts). The trajectories are simulated with the full timescale hierarchy ($\tau_a = 10^{-4}$, $\tau_W = 10^{-1}$, $\tau_b = 10^{2}$), so biases are free to evolve but do so much more slowly than the angles.

For \emph{same-sign readouts} ($\vec{a}^\star = (1, 1)^\T$, corresponding to teachers in the minimally dissimilar regime), the loss landscape has no local minima other than the global minimum: all gradient flow trajectories converge regardless of initialization. In contrast, for \emph{mixed-sign readouts} ($\vec{a}^\star = (-1, 1)^\T$, corresponding to teachers in the maximally dissimilar regime), the landscape develops local minima. In these configurations, both student neurons align to the same teacher neuron, leaving the other unmatched. The slow biases do not help escape these minima, confirming that they are robust features of the landscape rather than artifacts of the zero-bias constraint.

To verify that these findings are not specific to a single teacher geometry, we sweep the teacher angle $\theta^\star$ from $\pi/16$ to $\pi$ in 16 steps (Figure~\ref{fig:ana-combined}E). For each value of $\theta^\star$, we run gradient flow from a grid of random initializations and record the fraction that converge to a local (non-global) minimum. For same-sign readouts ($\vec{a}^\star = (1,1)^\T$), no trajectories are trapped at any angle, confirming the absence of local minima across the full range. For mixed-sign readouts ($\vec{a}^\star = (-1,1)^\T$), a nonzero fraction of trajectories is trapped at all tested angles: the basin of attraction of the local minima varies with $\theta^\star$, but the local minima persist throughout.

\section{Additional simulation results}

\subsection{Overall convergence statistics for the different settings}

\begin{longtable}{@{}llllrr@{}}
\caption{Overall convergence statistics across all experimental settings. All settings start with 1500 runs. The \emph{$D$ sweep} varies the input dimensionality $D \in \{2,4,8,16\}$ with the number of teacher nodes fixed at $M=4$; the \emph{$M$ sweep} varies the number of teacher nodes $M \in \{2,4,8,16\}$ with input dimensionality fixed at $D=2$. (We repeated the shared $M=4, D=2$ setting both times independently.) In all settings the student has the same number of nodes and the same activation function as the teacher ($\rho=1$). \emph{Maximal}\ = maximally dissimilar distribution; \emph{Minimal}\ = minimally dissimilar distribution.}
\label{stab:convergence-summary} \\
\toprule
Distribution & Activation & Sweep & Value & Completed & Global min \\
\midrule
\endfirsthead
\multicolumn{6}{l}{\small\itshape Continued from previous page} \\[2pt]
\toprule
Distribution & Activation & Sweep & Value & Completed & Global min \\
\midrule
\endhead
\midrule
\multicolumn{6}{r}{\small\itshape Continued on next page} \\
\endfoot
\bottomrule
\endlastfoot
\multirow{24}{*}{Maximal} & \multirow{8}{*}{ReLU} & \multirow{4}{*}{$D$ sweep} & 2  & 1500 & 973  \\
 &  &  & 4  & 1500 & 1208 \\
 &  &  & 8  & 1500 & 1231 \\
 &  &  & 16 & 1000 & 725  \\
\cmidrule(l){3-6}
 &  & \multirow{4}{*}{$M$ sweep} & 2  & 1500 & 1076 \\
 &  &  & 4  & 1500 & 972  \\
 &  &  & 8  & 1500 & 757  \\
 &  &  & 16 & 267  & 8    \\
\cmidrule(l){2-6}
 & \multirow{8}{*}{Softplus} & \multirow{4}{*}{$D$ sweep} & 2  & 1500 & 1204 \\
 &  &  & 4  & 1500 & 1341 \\
 &  &  & 8  & 1500 & 1360 \\
 &  &  & 16 & 1003 & 875  \\
\cmidrule(l){3-6}
 &  & \multirow{4}{*}{$M$ sweep} & 2  & 1500 & 1333 \\
 &  &  & 4  & 1500 & 1204 \\
 &  &  & 8  & 1500 & 509  \\
 &  &  & 16 & 1172 & 0    \\
\cmidrule(l){2-6}
 & \multirow{8}{*}{Tanh} & \multirow{4}{*}{$D$ sweep} & 2  & 1500 & 1230 \\
 &  &  & 4  & 1500 & 1485 \\
 &  &  & 8  & 1500 & 1472 \\
 &  &  & 16 & 1001 & 961  \\
\cmidrule(l){3-6}
 &  & \multirow{4}{*}{$M$ sweep} & 2  & 1500 & 1491 \\
 &  &  & 4  & 1500 & 1230 \\
 &  &  & 8  & 312  & 167  \\
 &  &  & 16 & 65   & 0    \\
\midrule
\multirow{24}{*}{Minimal} & \multirow{8}{*}{ReLU} & \multirow{4}{*}{$D$ sweep} & 2  & 1500 & 462 \\
 &  &  & 4  & 1500 & 812 \\
 &  &  & 8  & 1500 & 951 \\
 &  &  & 16 & 1000 & 480 \\
\cmidrule(l){3-6}
 &  & \multirow{4}{*}{$M$ sweep} & 2  & 1500 & 784 \\
 &  &  & 4  & 1500 & 467 \\
 &  &  & 8  & 1500 & 114 \\
 &  &  & 16 & 302  & 3   \\
\cmidrule(l){2-6}
 & \multirow{8}{*}{Softplus} & \multirow{4}{*}{$D$ sweep} & 2  & 1500 & 142 \\
 &  &  & 4  & 1500 & 589 \\
 &  &  & 8  & 1500 & 503 \\
 &  &  & 16 & 1026 & 159 \\
\cmidrule(l){3-6}
 &  & \multirow{4}{*}{$M$ sweep} & 2  & 1500 & 816 \\
 &  &  & 4  & 1500 & 145 \\
 &  &  & 8  & 1500 & 0   \\
 &  &  & 16 & 1497 & 0   \\
\cmidrule(l){2-6}
 & \multirow{8}{*}{Tanh} & \multirow{4}{*}{$D$ sweep} & 2  & 1500 & 573  \\
 &  &  & 4  & 1500 & 1358 \\
 &  &  & 8  & 1500 & 1413 \\
 &  &  & 16 & 1000 & 950  \\
\cmidrule(l){3-6}
 &  & \multirow{4}{*}{$M$ sweep} & 2  & 1500 & 1423 \\
 &  &  & 4  & 1500 & 573  \\
 &  &  & 8  & 702  & 43   \\
 &  &  & 16 & 422  & 0    \\
\midrule
\multirow{24}{*}{Standard} & \multirow{8}{*}{ReLU} & \multirow{4}{*}{$D$ sweep} & 2  & 1500 & 727  \\
 &  &  & 4  & 1500 & 1107 \\
 &  &  & 8  & 1500 & 1178 \\
 &  &  & 16 & 1000 & 630  \\
\cmidrule(l){3-6}
 &  & \multirow{4}{*}{$M$ sweep} & 2  & 1500 & 1008 \\
 &  &  & 4  & 1500 & 725  \\
 &  &  & 8  & 1500 & 407  \\
 &  &  & 16 & 289  & 30   \\
\cmidrule(l){2-6}
 & \multirow{8}{*}{Softplus} & \multirow{4}{*}{$D$ sweep} & 2  & 1500 & 453  \\
 &  &  & 4  & 1500 & 988  \\
 &  &  & 8  & 1500 & 1161 \\
 &  &  & 16 & 1017 & 670  \\
\cmidrule(l){3-6}
 &  & \multirow{4}{*}{$M$ sweep} & 2  & 1500 & 1162 \\
 &  &  & 4  & 1500 & 457  \\
 &  &  & 8  & 1262 & 109  \\
 &  &  & 16 & 823  & 3    \\
\cmidrule(l){2-6}
 & \multirow{8}{*}{Tanh} & \multirow{4}{*}{$D$ sweep} & 2  & 1500 & 1001 \\
 &  &  & 4  & 1500 & 1421 \\
 &  &  & 8  & 1500 & 1464 \\
 &  &  & 16 & 1003 & 948  \\
\cmidrule(l){3-6}
 &  & \multirow{4}{*}{$M$ sweep} & 2  & 1500 & 1479 \\
 &  &  & 4  & 1500 & 1000 \\
 &  &  & 8  & 1129 & 259  \\
 &  &  & 16 & 432  & 2    \\
\end{longtable}

\subsection{The chosen loss threshold separates global from local minima}

\label{app:loss-threshold}

\begin{figure}\includegraphics[width=\textwidth]{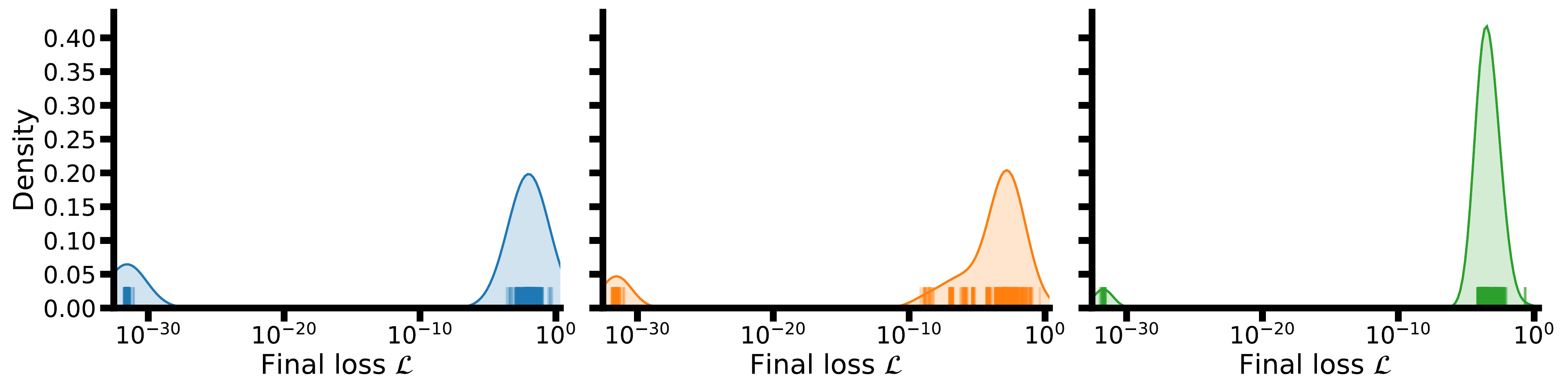}
    \caption{\textbf{Distribution of losses for converged student runs for all teacher networks with $M=4, D=2$.} (\textbf{A}) Distribution for teachers drawn from the maximal distribution. (\textbf{B}) Same as (A), but for teachers drawn from the standard distribution. (\textbf{C}) Same as (B), but for teachers drawn from the minimal distribution.}
    \label{sfig:loss-threshold}
\end{figure}

We use $1\times10^{-18}$ as the loss threshold to classify global from local minima in the main text. As shown in Figure~\ref{sfig:loss-threshold}, there is a large gap in losses below this threshold and those above for converged student runs. This indicates that the threshold is comfortably above the precision the system reaches, and that the higher loss values therefore do not correspond to global minima.

\subsection{Additional hyperparameter settings}

To ensure generalizability of the results, we run the simulations on additional settings. 

\begin{figure}\includegraphics[width=\textwidth]{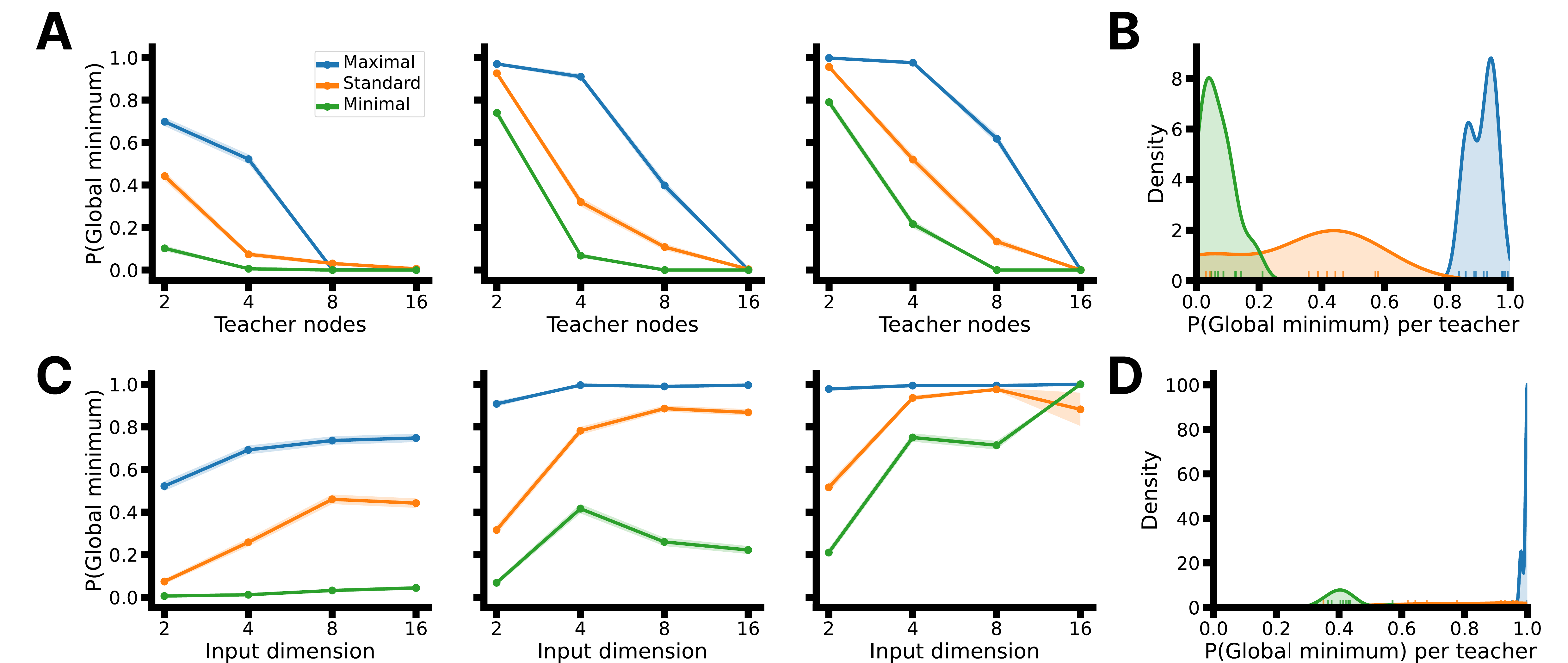}
    \caption{\textbf{Comparing the teachers reaching global minimum across all three distributions for the softplus activation function.} (\textbf{A}) Fraction of students reaching the global minimum as the number of teacher nodes are increased for $D=2$. The shaded region indicates the standard error of the binomial proportion.  (\textbf{B}) Frequency of students reaching the global minimum for teachers with $D=2$ and $M=4$, smoothed using Gaussian kernels. (\textbf{C}) Same as (A), but for increasing input dimensionality $D$ and for fixed number of teacher nodes $M=4$. (\textbf{D}) Same as (B), but for teachers with $D=4$ and $M=4$.}
    \label{sfig:softplus}
\end{figure}

\begin{figure}\includegraphics[width=\textwidth]{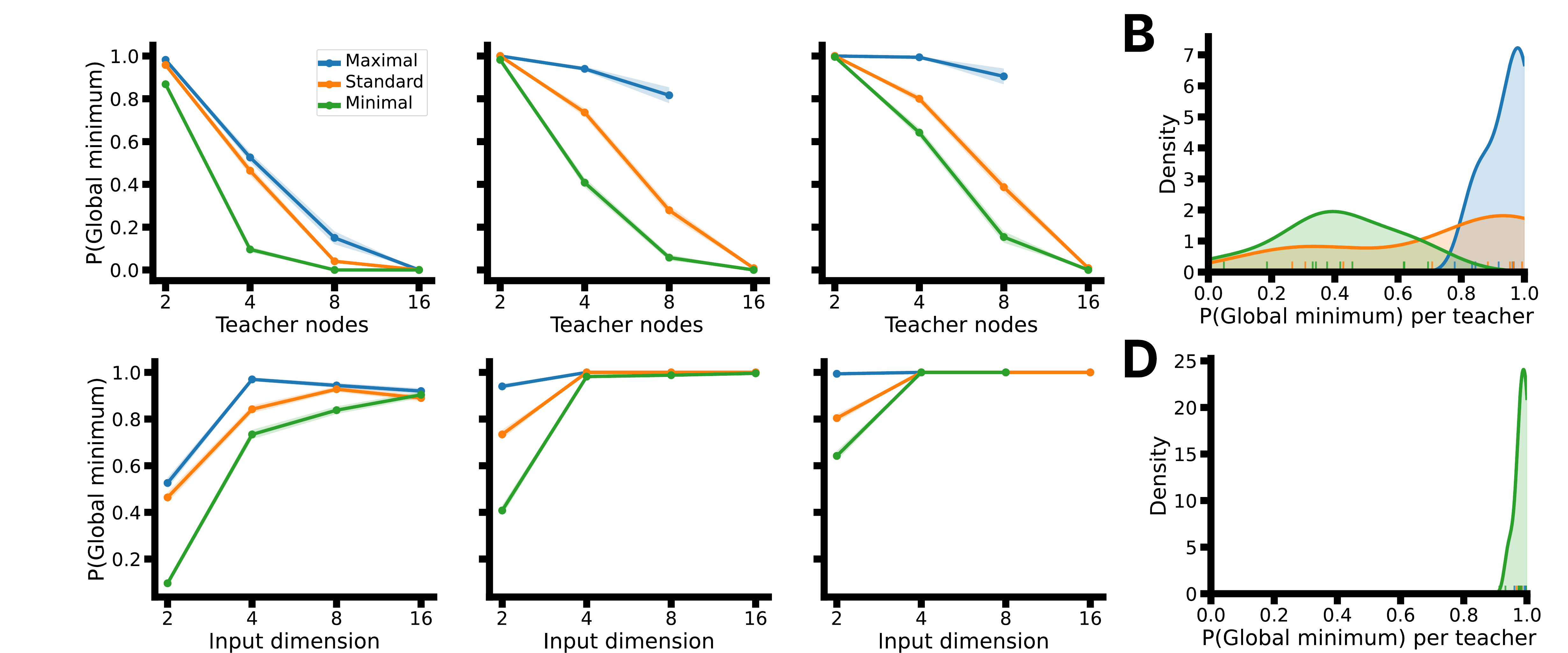}
    \caption{\textbf{Comparing the teachers reaching global minimum across all three distributions for the sigmoidal tanh activation function.} (\textbf{A}) Fraction of students reaching the global minimum as the number of teacher nodes are increased for $D=2$. The shaded region indicates the standard error of the binomial proportion.  (\textbf{B}) Frequency of students reaching the global minimum for teachers with $D=2$ and $M=4$, smoothed using Gaussian kernels. (\textbf{C}) Same as (A), but for increasing input dimensionality $D$ and for fixed number of teacher nodes $M=4$. (\textbf{D}) Same as (B), but for teachers with $D=4$ and $M=4$.}
    \label{sfig:tanh}
\end{figure}

\paragraph{Additional activation functions} We rerun all teacher sizes and input dimensionalities  across the three different teacher distributions for the softplus activation function in Figure~\ref{sfig:softplus} and the tanh activation function in Figure~\ref{sfig:tanh}. In both cases, the results are qualitatively very similar to those for ReLU.

\paragraph{Teacher distributions with different readout signs}

\begin{figure}
\centering
    \includegraphics[width=1\textwidth]{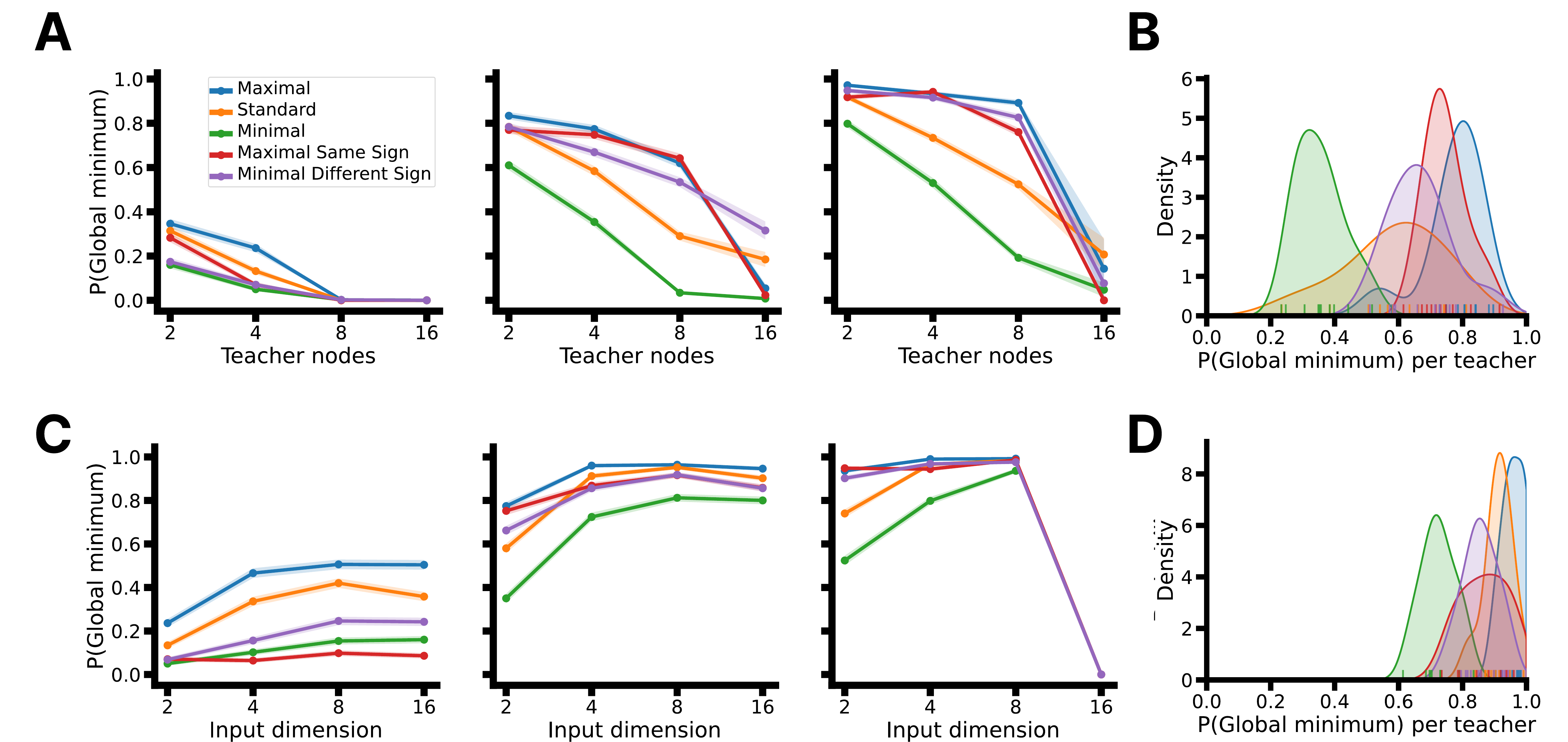}
    \caption{\textbf{Probability of reaching global minimum for teachers with different readout signs.} Error bars represent standard error of the binomial proportion.}
    \label{sfig:signs-sweep}
\end{figure}

\begin{figure}
\centering
    \includegraphics[width=0.5\textwidth]{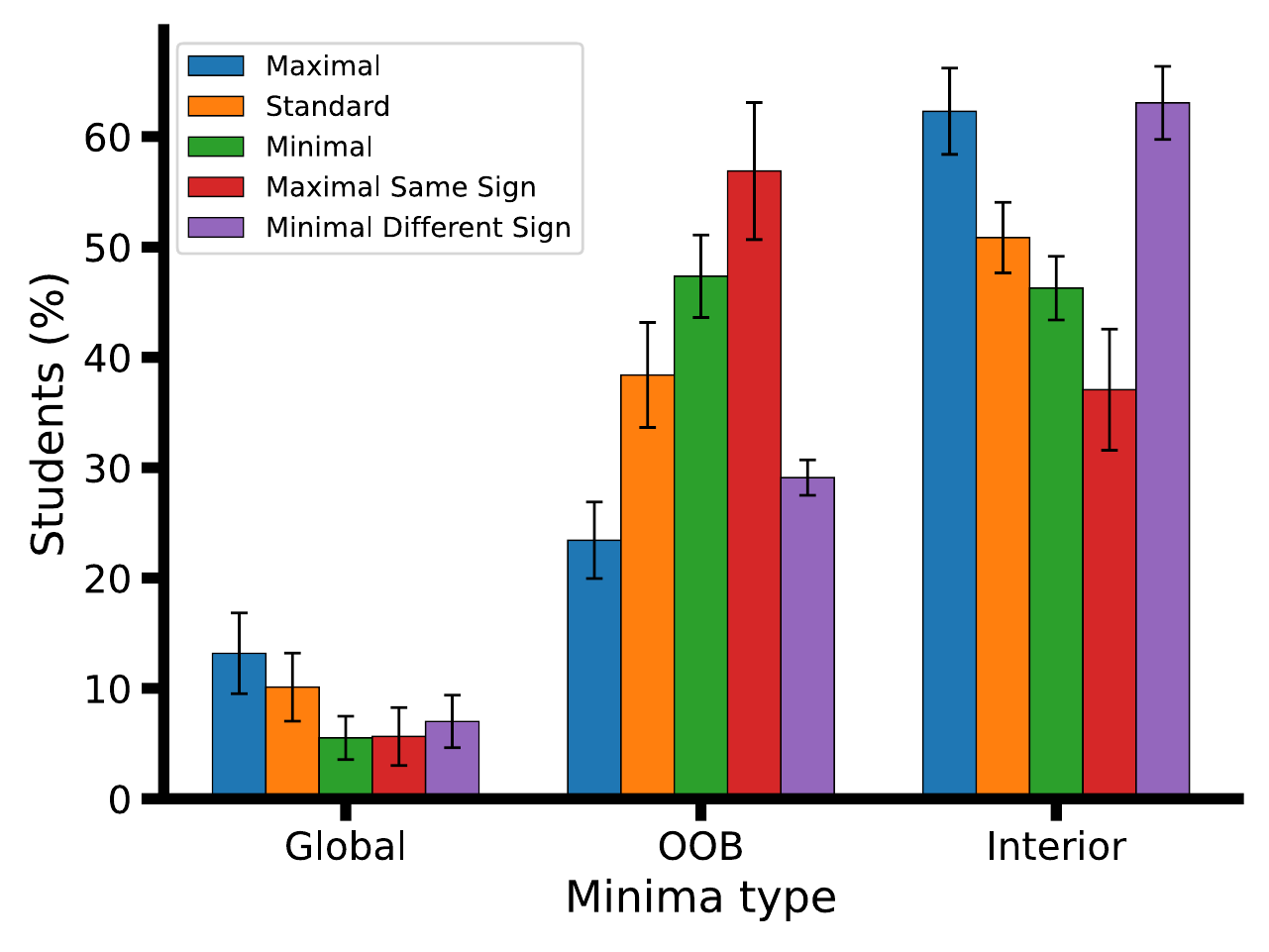}
    \caption{\textbf{Minima types for teachers with different readout signs aggregated across settings.} Error bars represent standard error of the binomial proportion.}
    \label{sfig:signs-minima}
\end{figure}

We also evaluated the students on maximally dissimilar teachers with the same readout signs, and minimally dissimilar teachers with different readout signs (Figures~\ref{sfig:signs-sweep} and~\ref{sfig:signs-minima}).

\begin{figure}
\centering
    \includegraphics[width=0.75\textwidth]{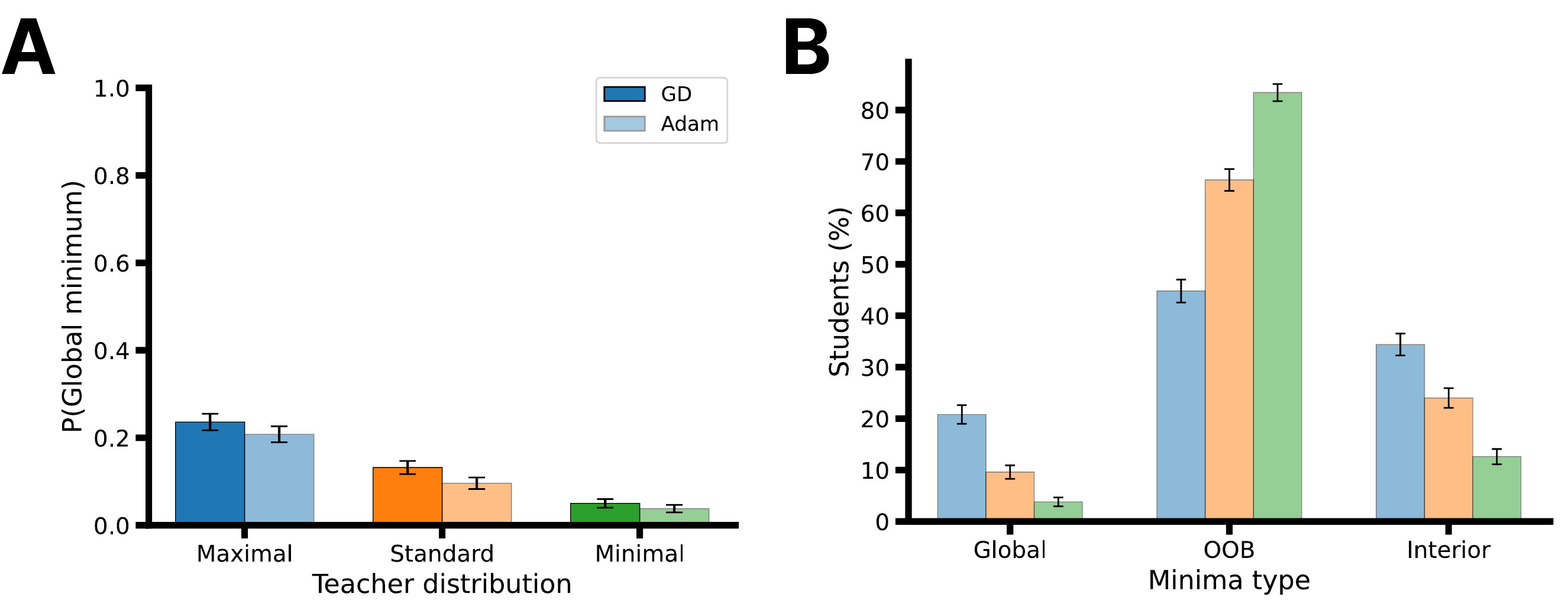}
    \caption{\textbf{Results for students optimized using Adam for $M=4$, $D=2$.} (\textbf{A}) Success results for students optimized with (\textit{light}) Adam, compared with (\textit{dark}) gradient flow for each of the three different teacher distributions. (\textbf{B}) Types of minima reached using Adam.  Error bars represent standard error of the binomial proportion.}
    \label{sfig:adam}
\end{figure}

\paragraph{Adam optimizer} We evaluated how the results change when the optimization is performed with Adam, rather than gradient flow (Figure~\ref{sfig:adam}). Again, the results are qualitatively similar.

\begin{figure}
\centering
    \includegraphics[width=0.75\textwidth]{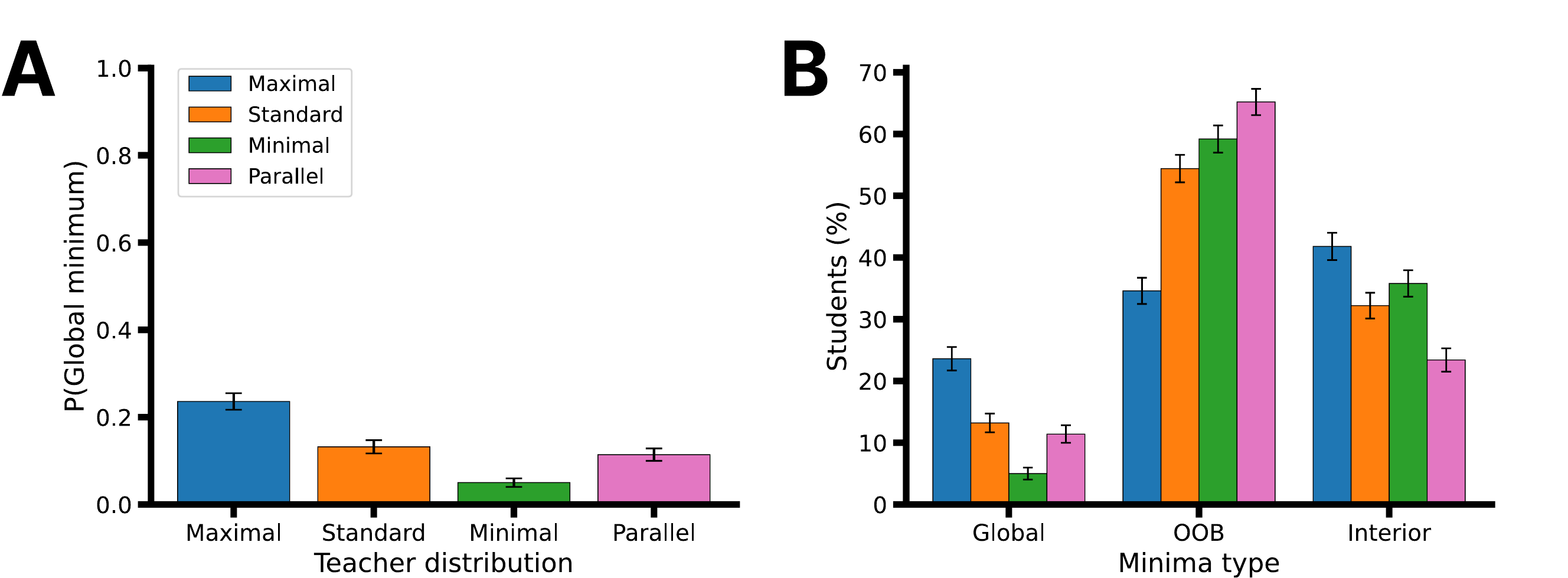}
    \caption{\textbf{Results for teachers with perfectly parallel hyperplanes.} (\textbf{A}) Success results for students on teachers drawn from (\textit{pink}) the distribution with parallel hyperplanes, compared with the other three teacher distributions. (\textbf{B}) Types of minima reached on teachers with parallel hyperplanes, compared with the other three distributions.  Error bars represent standard error of the binomial proportion.}
    \label{sfig:parallel}
\end{figure}

\paragraph{Parallel hyperplanes} 
We evaluated the students on a teacher distribution with parallel hyperplanes (Figure~\ref{sfig:parallel}).

\subsection{Empirical results for small networks}

\subsubsection{Single-node teacher-student systems}

We first ask how important starting similarity is between student and teacher network in determining success in finding the global minimum.

For this, we begin with targeted experiments in which we initialize a single teacher node ($M=1$) with

\begin{equation}
    \vec{w}^\star = \frac{1}{\sqrt{D}} (1, \dots, 1)^\T, \qquad b^\star = 0, \qquad a^\star = 1 .
\end{equation}

\begin{figure}
    \includegraphics[width=\textwidth]{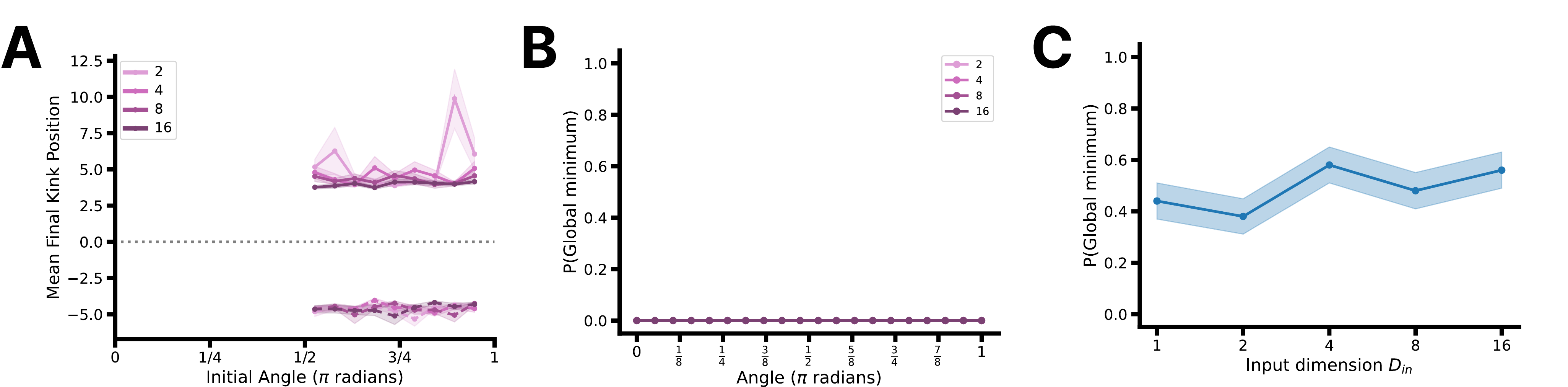}
    \caption{\textbf{Importance of initial similarity between student and teacher neuron in different number of input dimensions.} (\textbf{A}) Final kink position for neurons that reach a suboptimal local minimum in single-neuron student-teacher systems. Across (\textit{shade of purple}) dimensionalities, student nodes are driven out-of-bounds for distant initializations, corresponding to either a (\textit{solid}) positive or (\textit{negative}) kink. (\textbf{B}) Success rates for teachers and students with a single neuron when the student has an opposite sign of the teacher. The shaded region indicates the standard error of the binomial proportion. (\textbf{C}) Success rates for a single neuron as a function of input dimensionality. The shaded region indicates the standard error of the binomial proportion. (\textbf{D})  For a $D=2$ system with a single node, we systematically vary the kink. The shaded region indicates the standard error of the binomial proportion.}
    \label{sfig:single-neuron}
    \label{sfig:single-neuron-kink-diverged}
\end{figure}

We initialize single-node students ($N=1$) along a line on the sphere $\|\vec{w}\|_2 = 0.1$ with $b = 0$. For both values $a = 0.1$ and $a = -0.1$, we vary the angle $\theta$ between $\vec{w}$ and $\vec{w}^\star$ over 20 linearly spaced values between 0 and $\pi$. We repeat the experiment for dimensionalities in $\{1, 2, 4, 8, 16\}$. Figure~\ref{fig:ana-combined}B shows the convergence rate as a function of initial angle for same-sign student-teacher systems. Figure~\ref{sfig:single-neuron}A shows the final kink position as a function of initial angle. For systems where teachers and students were initialized to be opposite signs, none of the students were successful, regardless of starting angle (Figure~\ref{sfig:single-neuron}B).

If instead of constructing the student initializations in this way, we instead randomly sample the initial student weights according to the Glorot distribution, we see a slight positive trend in success rates as a function of dimensionality (Figure~\ref{sfig:single-neuron}C).

\paragraph{Varying the kink of a teacher neuron}

\begin{figure}
    \centering
    \includegraphics[width=0.5\textwidth]{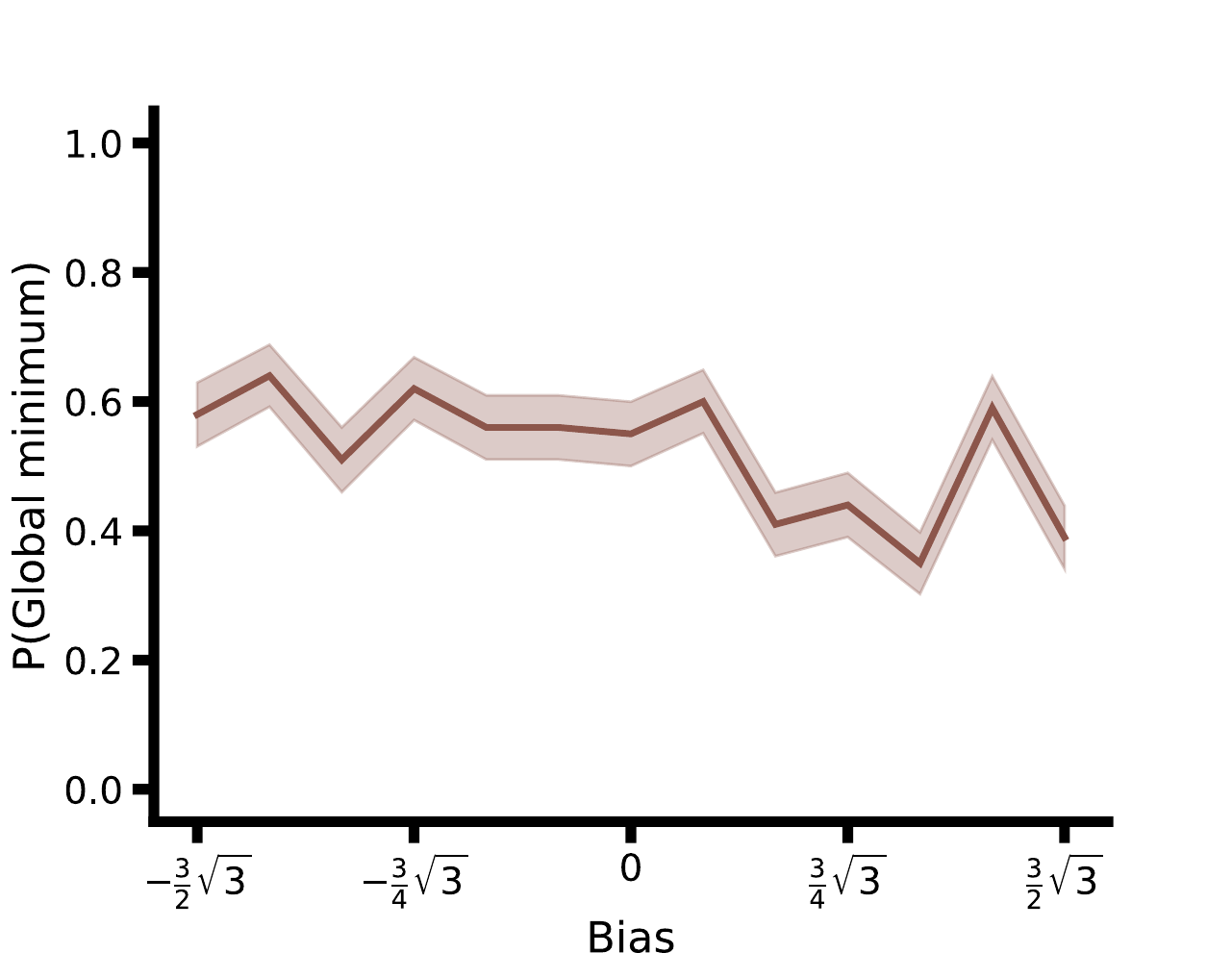}
    \caption{\textbf{Success rates for a single neuron as a function of the bias.} For a $D=2$ system with a single node, we systematically vary the bias, which changes the kink. The shaded region indicates the standard error of the binomial proportion.}
    \label{sfig:single-neuron-bias}
\end{figure}

We move a single node to different positions along the same line, finding that convergence rates are higher the earlier the ReLU's ``kink" is, corresponding to the amount of activated input space (Figure~\ref{sfig:single-neuron-bias}). In both cases, however, the differences between the settings are relatively slight when only considering a small number of nodes, meaning that the strong differences seen in Figure~\ref{fig:compare-losses} arise primarily from the interaction between different factors.

\subsection{Two-neuron teacher-student systems}

Next, we perform an experiment where we fix a teacher node, place a second teacher node with a given offset, and rotate it once around. We calculate the student success rate at each level. We find that the success rate roughly increases as the student node is rotated around to point in the exact opposite direction of the teacher node. Further, success rates are larger across the board when the two teacher nodes in the network have different signs than when they have the same sign (Figure~\ref{fig:ana-combined}F).

\subsection{Extended analysis of the local minima}

\begin{figure}
    \includegraphics[width=\textwidth]{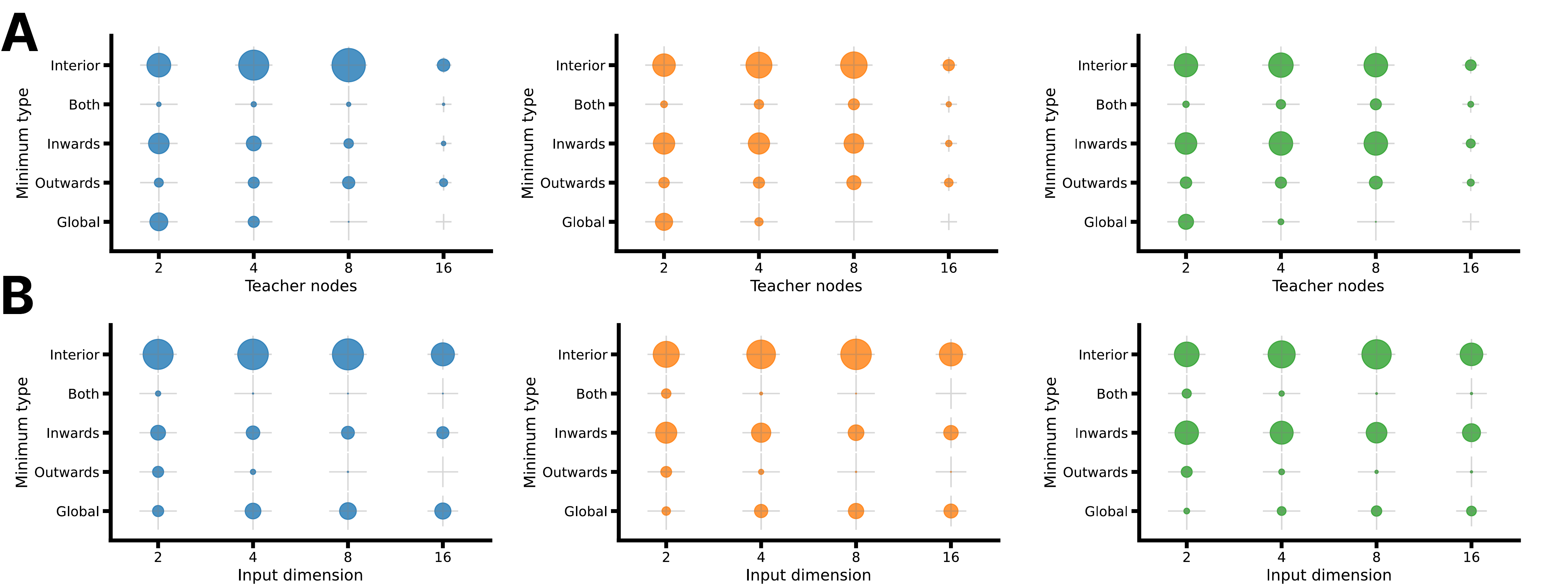}
    \caption{\textbf{Local minima reached by the system for $\rho=1$.} (\textbf{A}) Bubble plots showing how the reached minima change for increasing $M$ for each of the different teacher distributions with $D=2$ constant.  The size of the bubbles indicates how frequently that neuron is reached for each of the different teacher distributions. The crosses indicate the scale that corresponds to 100\% of neurons. (\textbf{B}) Same as in (A), but for increasing $D$ with $M=4$ constant.}
    \label{sfig:minima-types}
\end{figure}

\paragraph{Minima types} We plot the types of minima reached by the different teacher distributions at each level in Figure~\ref{sfig:minima-types}.

\begin{figure}
    \centering
    \includegraphics[width=0.4\textwidth]{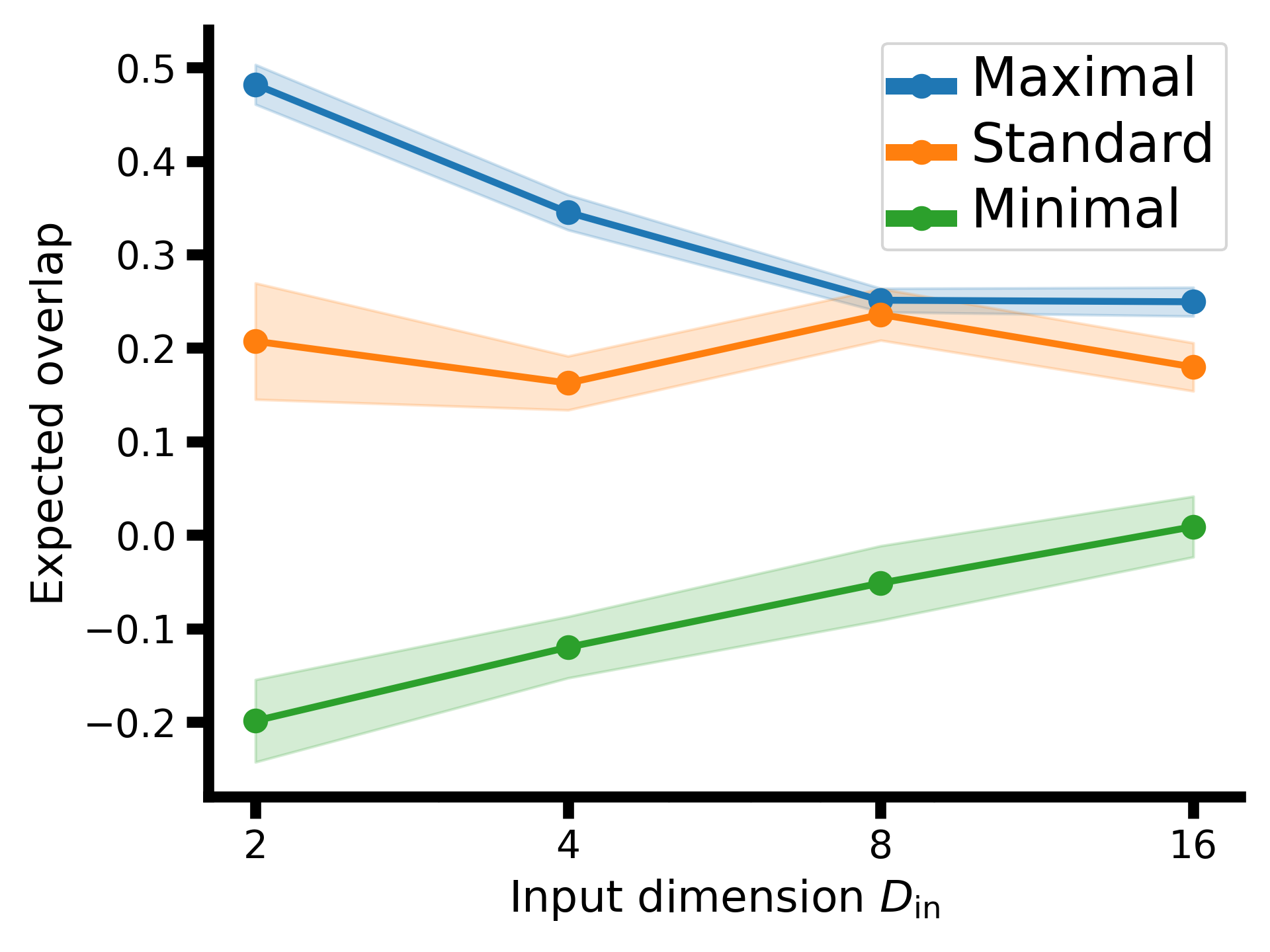}
    \caption{\textbf{Average starting overlap as a function of increasing dimensionality.}}
    \label{sfig:overlap-dimensionality}
\end{figure}

\paragraph{Overlap terms} Overlap terms tend towards zero for increasing dimensionality (Figure~\ref{sfig:overlap-dimensionality}). 

\begin{figure}
    \centering
    \includegraphics[width=\textwidth]{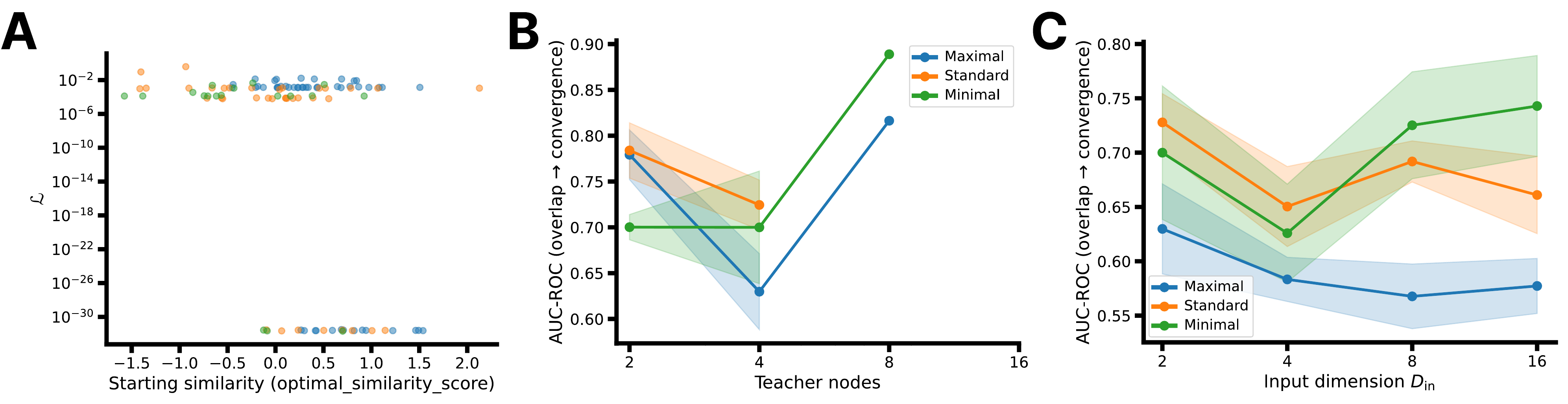}
    \caption{\textbf{Relationship between starting similarity and final success, as quantified using AUROC.}}
    \label{sfig:auroc}
\end{figure}

\paragraph{AUROC prediction} We plot individual overlaps vs. losses for teachers for a single setting in Figure~\ref{sfig:auroc}.

\pagebreak

\end{document}